\documentclass{article} % For LaTeX2e
\usepackage{iclr2026_conference,times}

\usepackage{amsmath,amsfonts,bm}

\def\eqref#1{equation~\ref{#1}}
\def\1{\bm{1}}

\DeclareMathAlphabet{\mathsfit}{\encodingdefault}{\sfdefault}{m}{sl}
\SetMathAlphabet{\mathsfit}{bold}{\encodingdefault}{\sfdefault}{bx}{n}

\usepackage{colortbl}
\usepackage{hyperref}
\usepackage{url}
\usepackage{booktabs}
\usepackage{tabularx}
\usepackage{amssymb}
\usepackage{arydshln}
\usepackage{multirow}
\usepackage{graphicx}
\usepackage[svgnames]{xcolor}
\usepackage[misc]{ifsym}
\usepackage{wrapfig}
\usepackage{subcaption}
\usepackage{enumitem}
\usepackage{algorithm}
\usepackage{algpseudocode}
\usepackage{fontawesome}
\usepackage{hyperref}

\title{ProxyAttn: Guided Sparse Attention via Representative Heads}

\author{
Yixuan Wang\textsuperscript{\rm 1,}\footnotemark[1],\quad
Huang He\textsuperscript{\rm 2},\quad
Siqi Bao\textsuperscript{\rm 2},\quad
Hua Wu\textsuperscript{\rm 2},\quad
Haifeng Wang\textsuperscript{\rm 2},\\
\textbf{
Qingfu Zhu\textsuperscript{\rm 1},\quad
Wanxiang Che\textsuperscript{\rm 1,}\footnotemark[2]
}
\\
\\
\textsuperscript{\rm 1}Harbin Institute of Technology, China \\
\textsuperscript{\rm 2}Baidu Inc., Beijing, China \\
\texttt{\{yixuanwang, qfzhu, car\}@ir.hit.edu.cn} \\
\texttt{\{hehuang, baosiqi, wu\_hua, wanghaifeng\}@baidu.com} \\
}

\iclrfinalcopy % Uncomment for camera-ready version, but NOT for submission.

\begin{document}

\maketitle

\renewcommand{\thefootnote}{\fnsymbol{footnote}}
\footnotetext[1]{Work done during internship at Baidu, Inc.}
\footnotetext[2]{Corresponding author.}
\renewcommand{\thefootnote}{\arabic{footnote}}

\begin{abstract}
% BG
The quadratic complexity of attention mechanisms limits the efficiency of Large Language Models (LLMs) on long-text tasks.
% Sparse
Recently, methods that dynamically estimate block importance
have enabled efficient block sparse attention,
leading to significant acceleration in long-text pre-filling of LLMs.
% Problems
However, their coarse-grained estimation inevitably leads to performance degradation at high sparsity rates.
% Solution
In this work, we propose \textit{ProxyAttn},
a training-free sparse attention algorithm
that achieves more precise block estimation by compressing the dimension of attention heads.
% Observation
Based on our observation of the similarity among multiple attention heads,
we use the scores of pooled representative heads to approximate the scores for all heads.
% Budget
To account for the varying sparsity among heads, we also propose a block-aware dynamic budget estimation method.
% Result
By combining the scores from representative proxy heads with multi-head dynamic budgets,
we achieve a more fine-grained block importance evaluation at low computational cost.
% Experiment
Experiments on a variety of mainstream models and extensive benchmarks
confirm the underlying similarity among attention heads.
% Conclusion
Leveraging a fine-grained estimation,
the proposed method achieves substantial gains in performance and efficiency compared to existing methods.
% Specific
More precisely, ProxyAttn can
achieve up to 10.3x attention acceleration and 2.4x prefilling acceleration
without significant performance loss.
Our code is available at \faGithubAlt~\url{https://github.com/wyxstriker/ProxyAttn}.
\end{abstract}

\section{Introduction}
% LLMs have achieved great success
Large Language Models (LLMs) have shown outstanding performance in long context tasks
\citep{liu2025comprehensive,liu2025survey}
and are widely applied in fields such as code generation \citep{jimenez2023swe,hui2024qwen2}
and mathematical reasoning \citep{chen2025towards,guo2025deepseek}.
% Sufferred from long input text
However, transformer-based LLMs face efficiency limitations
when processing long texts with millions of tokens.
% due to attention's quadratic complexity
The quadratic complexity of the attention mechanism poses a significant challenge to the inference of models.
% Reasearch start
Recently, numerous studies \citep{brauwers2021general,zhang2025efficient} have worked on
refining the attention module to increase the training and inference efficiency.

% To solve this problem, SparseAttenion
In particular, sparse attention methods mitigate the latency issues of the attention mechanism
by skipping certain computations.
% Sparse attention methods
By leveraging the inherent sparsity \citep{liu2022dynamic,deng2024attention} of attention score distributions,
these methods can reduce computational costs while preserving model performance.
% Pattern Method
Some static methods \citep{beltagy2020longformer,zaheer2020big} use heuristic fixed templates to achieve sparse computation,
making it difficult to maintain model performance with diverse attention patterns \citep{likhosherstov2021expressive}.
% Dynamic Method
To achieve a more accurate estimation,
many efforts have focused on exploring
various low-cost method to assess the importance of tokens,
which in turn enables the dynamic sparse patterns.
% Block-Spase-Attention
The performance of hardware-friendly sparse attention,
which is typically handled at the block level \citep{guo2024blocksparse},
is heavily dependent on the efficiency and accuracy of its importance estimation.
% MInference
MInference \citep{jiang2024minference} leverages multiple pre-defined templates and an online computed index
to guide the construction of block sparse patterns.
% FlexPrefill
Furthermore, FlexPrefill \citep{lai2025flexprefill} proposes an input-aware approach
that dynamically determines the sparse patterns and sparsity rates.

% Current Problem
While existing methods have proven effective in long-context scenarios,
their performance at high sparsity rates is still
\textbf{limited by their coarse-grained importance estimation}. 
% current solutions from seq-level compression
To evaluate the importance of each block,
these methods typically compress along the sequence dimension to approximate their attention scores.
% Weakness
This coarse-grained compression can cause a few high-scoring positions to be overlooked,
which in turn impacts the final performance.
% token-wise
A token-level dot-product calculation would solve this issue but is not feasible,
as its time complexity scales identically to full attention.

% what we do? from head view
In this paper, we achieve a fine-grained and efficient trade-off in block importance estimation
by exploring compression of the attention head dimension.
% Observation
We begin by analyzing the performance of multi-head attention on long sequences,
where we observed two primary characteristics of how the heads behave:
\textbf{(1) The overall attention trends of different heads are consistent.}
\textbf{(2) The primary differences between heads lie in their sparsity.}
% Method ProxyAttn
Based on these findings, we propose \textit{ProxyAttn},
a training-free sparse attention algorithm that
approximates attention scores using a few representative heads.
% Specifically
Specifically, we leverage the similarity among attention heads
to approximate the full attention scores using the complete scores from a few pooled heads.
% MaxPooling
By incorporating max pooling, our proposed method enables more accurate block importance estimation while ensuring efficiency.
% Dynamic Budget
Furthermore, given the significant differences in head sparsity,
we propose an online block-aware budget estimation method.
% Combination
By integrating the unified importance score with head-level budget,
we can achieve diverse sparse attention patterns across different heads.

\begin{figure}[t]
   \centering
   \includegraphics[width=\textwidth]{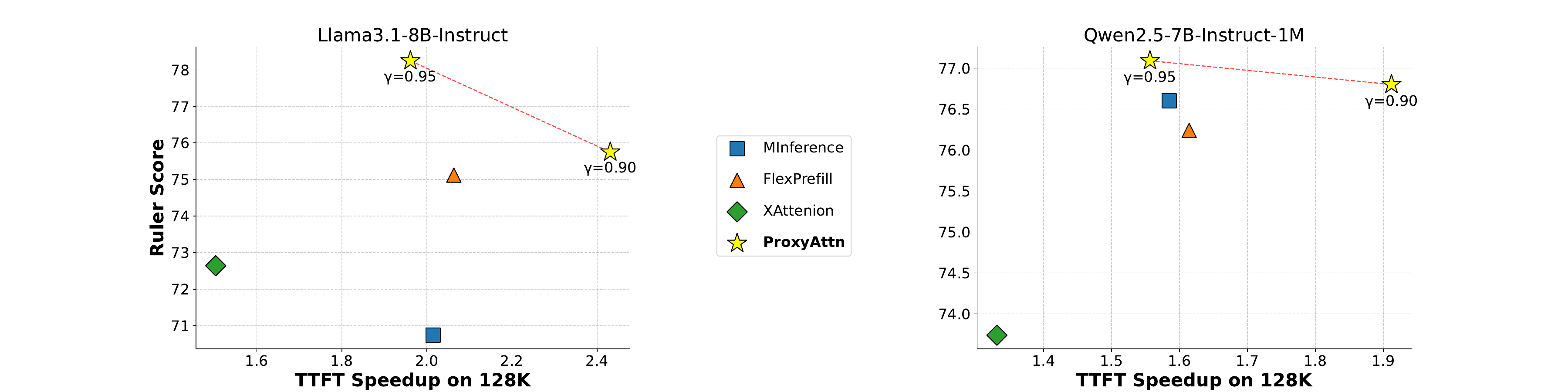}
   \caption{Performance and speedup results of different sparse attention methods on RULER.}
   \label{fig:intro}
\end{figure}

% backbone
In order to evaluate our proposed method, we conduct a series of extensive experiments.
We use widely-adopted LLMs that are specifically designed for long-context tasks
like Llama3.1-8B-Instruct \citep{grattafiori2024llama} and Qwen2.5-7B-Instruct-1M \citep{yang2025qwen2},
along with a variety of mainstream benchmarks
such as RULER \citep{hsieh2024ruler}, InfiniteBench \citep{zhang2024bench}, and LongBench-v2 \citep{bai2024longbench,bai2024longbench2}.
% Result
Experimental results show that ProxyAttn significantly outperforms other methods on both synthetic and real-world datasets.
% Trade-off
% By using a more fine-grained importance assessment,
% our proposed method achieves better performance at a similar sparsity rate
% and offers greater acceleration while maintaining performance.
% Figure
As illustrated in Figure \ref{fig:intro},
the proposed ProxyAttn achieves a Pareto front for performance and efficiency compared to other coarse-grained estimation methods.
Our contributions can be summarized as follows:
\begin{itemize}[leftmargin=2em]
\item We introduce ProxyAttn, a training-free sparse attention estimation method.
It leverages the attention scores from a small number of proxy heads to efficiently estimate the scores for all heads,
achieving more accurate importance measures.
\item Considering the diverse sparsity among heads, we propose a block-aware budget allocation method.
Coupled with the unified importance score,
our method can provide different head sparsity budgets online,
which in turn yields diverse sparse attention masks.
\item Extensive experiments demonstrate that the proposed ProxyAttn
outperforms other coarse-grained sparse attention methods
across various model architectures and benchmarks.
Our approach yields a maximum 10.3x speedup for attention computation
% and a 2.4x end-to-end prefill speedup,
without significant performance loss.

\end{itemize}

\section{Observation}

\subsection{Motivation}
% Block Sparse Attention
Recent work \citep{jiang2024minference,lai2025flexprefill,gao2024seerattention}
on the importance of different attention blocks typically
employs heuristic ``vertical and slash" patterns or pooling methods along the sequence dimension for approximation.
% Problems
These methods either rely on specific heuristic template assumptions
or opt for a coarse-grained estimation, 
rendering the importance measures of blocks inaccurate \citep{xu2025xattention}.
% % Block Level
% For an input sequence of length $N$ and a block size $b$,
% the block importance calculation for head $i$ using a pooling method
% can be formalized as:
% % $$A = Softmax(\frac{Pooling(Q)Pooling(K)^T}{\sqrt{d}}), A \in \mathbb{R}^{\frac{N}{b} \times \frac{N}{b}}$$
% \begin{equation}
%    \mathbf{A}_i = \text{softmax}\left(\frac{\text{avgpool}(\mathbf{Q}_i)\text{avgpool}(\mathbf{K}_i)^T}{\sqrt{d_k}}\right)
% \end{equation}
% where $\mathbf{Q}_i, \mathbf{K}_i \in \mathbb{R}^{N \times d_k}$ and $\mathbf{A}_i \in \mathbb{R}^{\frac{N}{b} \times \frac{N}{b}}$.
% Ignore
Given the inherent sparsity of attention distributions,
such coarse-grained pooling can easily overlook a few high-scoring positions.

% Our motivation
A viable approach for both accuracy and efficiency in token-level dot-product computations
is to compress along the head num dimension instead of the sequence dimension.
% more
This allows a few key heads to serve as proxies for estimating the importance of all heads.
% Condition
However, the feasibility of this approach hinges on the assumption that
attention heads exhibit a certain degree of consistency in their scores.
% What we do
To validate this hypothesis,
we conduct some observational studies
on the behaviors of multiple heads in a long-context setting.

\begin{figure}[t]
   \centering
   \subcaptionbox{The tokens that attention heads focus on
exhibit a consistency in long context.
   \label{fig:obs:same}}
   {\includegraphics[width=0.32\textwidth]{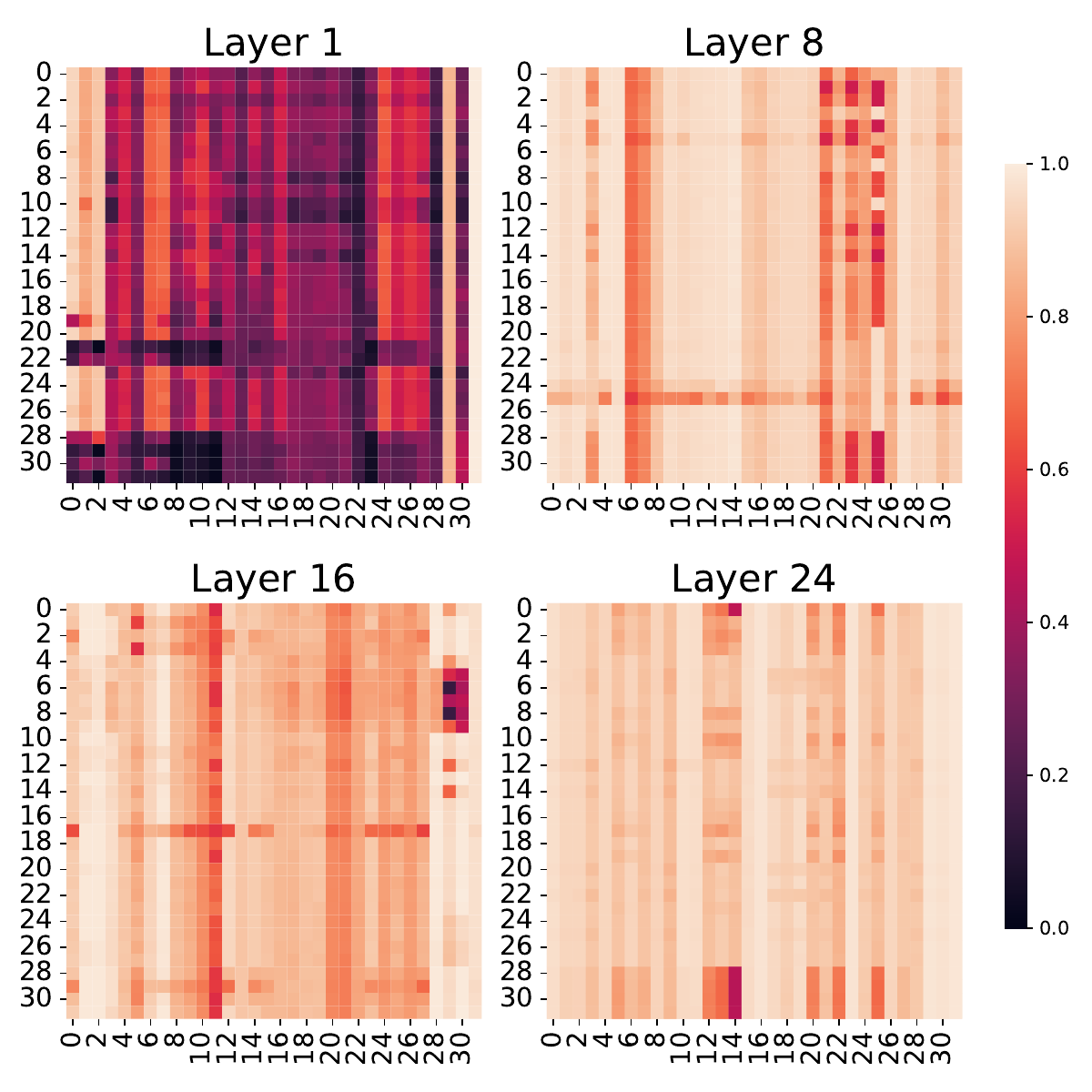}}
   \hfill
   \subcaptionbox{A shared token ranking can simultaneously
accommodate the scores from different heads.
   \label{fig:obs:actual}}
   {\includegraphics[width=0.31\textwidth]{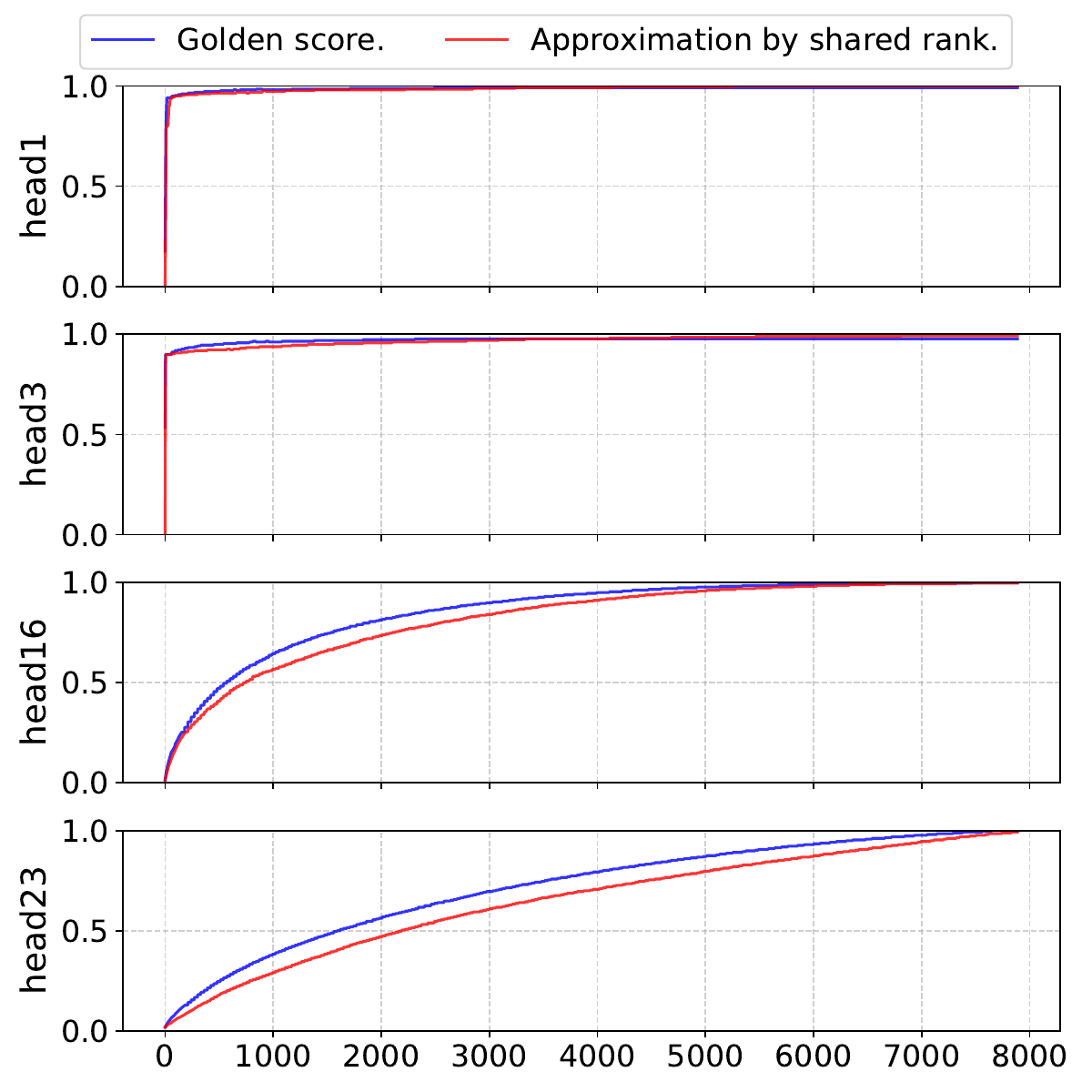}}
   \hfill
   \subcaptionbox{Attention heads vary in their sparsity.
Both sparse and dense heads still share a common focus on certain tokens.
   \label{fig:obs:diff}}
   {\includegraphics[width=0.31\textwidth]{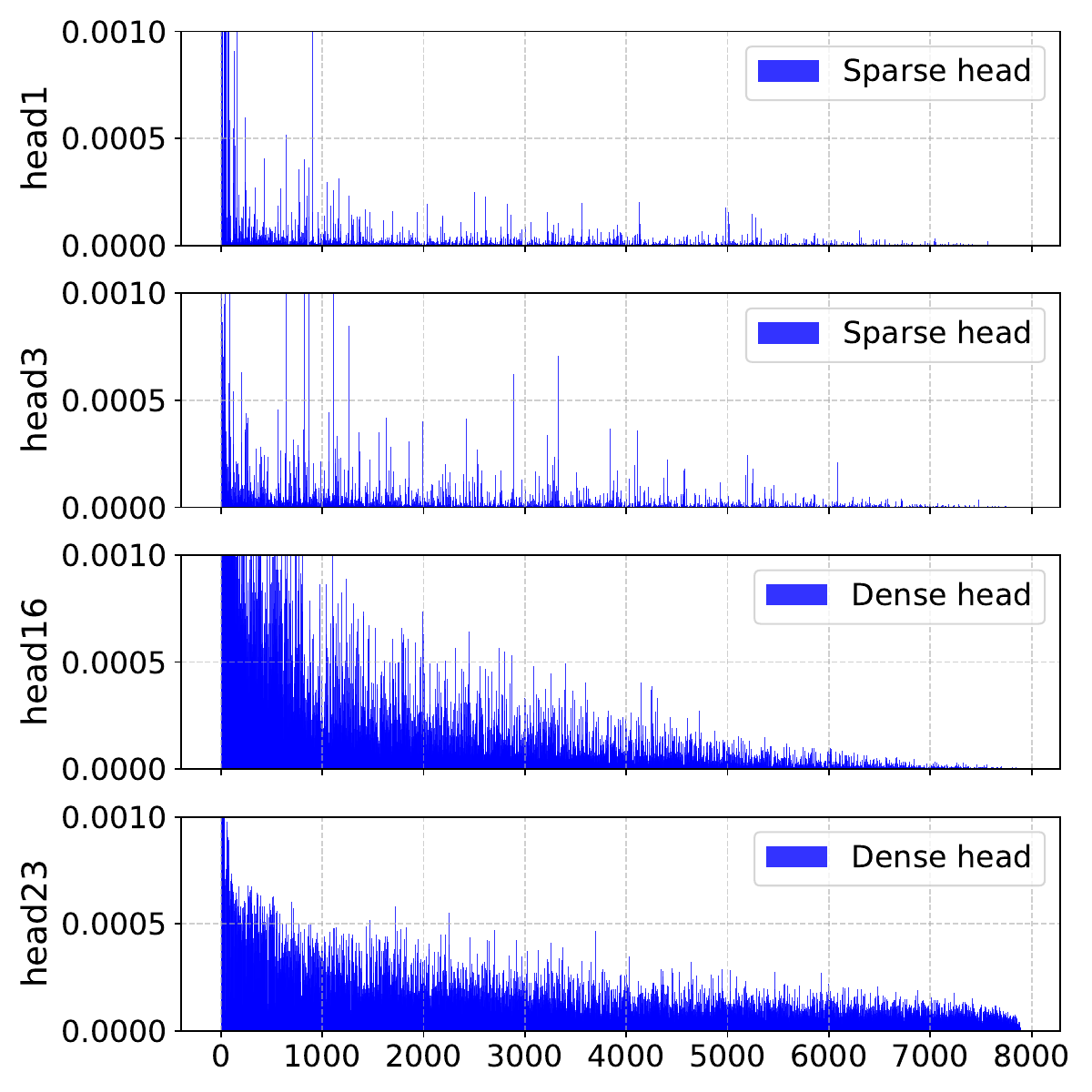}}
   \caption{Observational study on Llama3.1-8B-Instruct using 8K-token data
   from the Needle in a Haystack (NIAH) Single-1 task of the RULER \citep{hsieh2024ruler} dataset.
}
   \label{fig:obs}
\vspace{-1em}
\end{figure}

\subsection{Characteristics of Multiple Attention Heads}
\label{sec:finding}
% Setting
To investigate the similarities and differences among different heads,
we conduct experiments on the Llama model
using long-context data containing 8K tokens.
% Findings
As shown in Figure \ref{fig:obs}, by observing the performance of different heads,
we can derive two key findings:

\paragraph{Attention heads exhibit consistency in token focus.}
% Overlap Coverage
We first calculate the overlap of tokens that different heads primarily focus on.
As shown in Figure \ref{fig:obs:same},
the score at position (i, j) represents the cumulative attention score at head $j$
obtained by the top 1024 tokens from head $i$.
% Observation
It can be observed that while the very dense first layer has inherently low recall,
the attention overlap scores between different heads in other layers are very high.
% Conculsion1
This suggests that in long texts, multiple attention heads,
particularly in deeper layers, focus on a large intersection of tokens.

% Motivation
To further assess this similarity,
we consider whether it could be captured by a shared metric,
with a natural candidate being average pooling across the heads.
% PDF
As depicted in Figure \ref{fig:obs:actual},
we sort the tokens by their average score across all heads
and then analyze the discrepancy
between the cumulative probability curve based on a shared ranking
and the actual score curve for specific heads.
% Conclusions
The consistency between the two curves
serves as the basis for our ProxyAttn algorithm,
which leverages the scores from a few heads to approximate the attention of all heads.

\paragraph{Attention heads exhibit variability in sparsity.}
% Intro
While the diversity among attention heads is a well-established observation \citep{jiang2024minference},
we find that it does not contradict our findings of consistency.
% Observations
Consistent with the preceding setup,
we utilize the same shared token ranking to examine the attention scores across various heads.
% Figures
Figure \ref{fig:obs:diff} illustrates that the primary variation between heads is in their sparsity rather than the tokens they focus on.
% Same
This observation corroborates the findings of
\citet{xiao2024duoattention,fu2024moa}.
% Analysis
As seen in the figure, given the same ranking,
all heads assign high attention to the leading tokens.
% Difference
The main distinction is that some heads are very sparse and only attend to the initial portion,
while others are dense and maintain significant scores for later tokens.
% Conclusion
This motivates us to realize that while leveraging shared scores,
a flexible budget must be allocated to different heads to ensure performance.

\vspace{-0.5em}
\section{ProxyAttn}
\begin{figure}[t]
   \centering
   \includegraphics[width=0.9\textwidth]{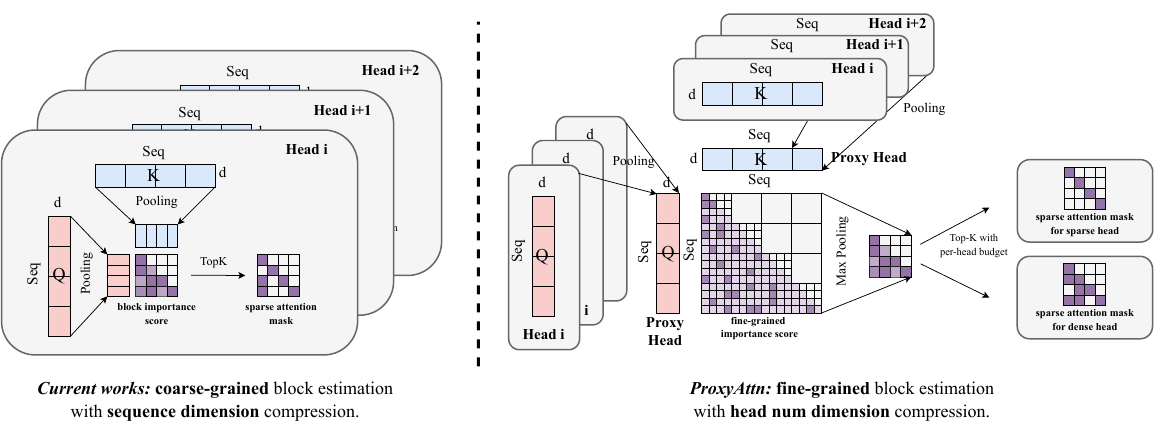}
   \caption{\textbf{Illustration of ProxyAttn.}
By compressing the head dimension, ProxyAttn can obtain token-level importance scores, leading to more accurate block importance estimation.
A proxy head is able to obtain diverse masks by leveraging the online budget estimations from different heads.
}
   \label{fig:arch}
   \vspace{-1em}
\end{figure}

% Overall
Building upon our above observations of attention heads,
we propose the ProxyAttn algorithm to efficiently achieve accurate, fine-grained importance estimation
for block sparse attention.
% Figure
A schematic comparison with existing coarse-grained estimation methods
is shown in Figure \ref{fig:arch}.
% Sec1
Leveraging the consistency of token focus among heads,
we first employ an intra-group score sharing method (\S \ref{method_pooling})
for importance estimation.
% Sec2
To account for the varying sparsity of heads,
we then dynamically assign a unique budget (\S \ref{method_budget})
to each head to generate diverse sparse masks.

\subsection{Unified Score Estimation}
\label{method_pooling}
% Motivation
In order to obtain the token-level dot-product scores for each block in the attention map,
compression must be applied along dimensions other than the sequence length to ensure efficiency.
% Seq & headdim
Given the consistency of token attention across heads on long contexts,
we propose to perform compression along the head num dimension.
% Specifically
Specifically, we partition the attention heads into distinct groups.
% meta
Within each group, a designated head is used to compute attention,
which then serves as a proxy to guide the attention scores for the entire group.
% Pooling
The entire process can be formulated as:
\begin{equation}
   \mathbf{A}^i = \text{maxpool}\left(\text{softmax}\left(\frac{\mathbf{Q}^g{\mathbf{K}^g}^T}{\sqrt{d_k}}\right)\right), \quad
   i \in G_g
\end{equation}
where $\mathbf{Q}^g, \mathbf{K}^g \in \mathbb{R}^{N \times d_k}$ and $\mathbf{A}_i \in \mathbb{R}^{\frac{N}{b} \times \frac{N}{b}}$.
For all heads belonging to group $g$,
a single token-level score will be shared.
% Effort
Combined with max-pooling, it enables fine-grained block importance estimation at the cost of computing only $|G|$ proxy heads.

% G AvgPooling
For the representative head queries $\mathbf{Q}^g$ and keys $\mathbf{K}^g$
, we heuristically apply the same average pooling as for the sequence.
% formated
This can also be seen as a form of compression along the head number dimension
and can be formalized as:
\begin{equation}
\mathbf{Q}^g = \frac{1}{|G|}\sum_{i \in G} \mathbf{Q}^i
\quad
\mathbf{K}^g = \frac{1}{|G|}\sum_{i \in G} \mathbf{K}^i
\end{equation}
% GQA
It should be noted that for models using Group-Query Attention (GQA) \citep{ainslie2023gqa},
the group granularity will be aligned with the keys, 
ensuring that all queries belonging to the same group of keys are also within that same group.

% Striker
To further reduce the computational cost of the estimation operation, we also dropped certain qk pairs by stride,
keeping only the first token in each stride window.
% Local
Driven by the local similarity inherent in attention mechanisms \citep{wu2024tokenselect},
this operation reduces the computational cost without significant loss of accuracy
(see Appendix \ref{app:stride} for a detailed analysis).
% Speed
For a model with $n$ heads,
the theoretical computational cost of using $g$ representative proxy heads
for estimation is expected to be
$\frac{g}{n \cdot \text{stride}^2}$ of full multi-head attention.
% Act
Leveraging head sharing,
the proposed method reduces the computational cost
for a 7B model to roughly one percent of full attention
under conventional settings.
% Overall
This allows for fine-grained estimation without compromising computational efficiency.

\subsection{Dynamic Budget Allocation}
\label{method_budget}
% Past Search
After obtaining importance scores for different blocks,
a common practice is to further obtain a sparse block mask \citep{gao2024seerattention,xu2025xattention},
often by using methods like Top-K selection.
% For our
However, directly applying this approach to proposed method 
would result in identical masks for all heads in the same group,
thereby impacting the maximum achievable sparsity rate.
% Last Block
Given the findings from \S \ref{sec:finding},
the attention focuses of different heads are consistent,
with the primary difference being in their sparsity.
% Mehtod
We consider an online approach to approximately evaluate the sparsity of individual heads,
which in turn guides the selection of the budget.

% block-aware budget allocation
Inspired by the work of \citet{jiang2024minference},
we propose to use the queries from the last block
to estimate the sparsity of different heads.
% define
The budget required for a given head is approximated
as the minimum budget for the query of the final block
to achieve a cumulative probability of $\gamma$.
% Block-aware
Furthermore, given the significant discrepancy between token-level computation
and the actual block budget due to attention sparsity,
we employ average pooling to ensure that the budget estimation is performed at the block level.
% alg
Algorithm \ref{alg:budget} presents the overall approach to budget estimation.
% what
After obtaining the budget $\mathbf{b}_i$ for each head,
we select the top $b_i$ blocks with the highest unified scores to form the sparse attention masks.
% TopK
The final sparse attention mask $\mathbf{S}^i_{M \times N}$ for head $i$ can be formalized as:
\begin{equation}
   \begin{aligned}
      \mathbf{S}^{i}_{mn} = \begin{cases}
          1, & \text{if } n \in \text{TopK}(\mathbf{A}^i[m], \text{K}=b_iN).\text{index} \\
          0, & \text{otherwise}
      \end{cases}
   \end{aligned}
\end{equation}

Where $\mathbf{M,N}$ is the block-level sequence length,
and $\mathbf{A}^i$ is the previously estimated block-level importance of head $i$.
% Conclusion
As shown in Figure \ref{fig:arch},
the proposed ProxyAttn achieves diverse sparse attention masks
by combining shared head scores and the dynamic budget allocation method.
% block sparse attention
By leveraging the efficient block sparse attention kernel \citep{guo2024blocksparse},
we can substantially accelerate the attention operation for long contexts
while maintaining comparable model performance.

\begin{algorithm}[t]
\caption{Block-aware Budget Estimation for Head $i$}
\label{alg:budget}
\begin{algorithmic}
\Require Query states $\mathbf{Q}^i$, key states $\mathbf{K}^i$, cumulative probability threshold $\gamma$
\vspace{0.1em}
\Ensure Estimated budget $b_i$
\State $\mathbf{\hat{A}}^i \gets \mathrm{softmax}\!\left(\mathbf{Q}^i_{\text{last}} {\mathbf{K}^i}^\top \,/\, \sqrt{d_k} \right)$ 
\Comment{Compute approximate attention scores}
\State $\mathbf{a}^i \gets \mathrm{avgpool}(\mathbf{\hat{A}}^i)$
\Comment{Aggregate into block-level scores}
\vspace{0.3em}
\State $\mathbf{a}^i \gets \mathrm{sort}(\mathbf{a}^i) \,/\, \mathrm{sum}(\mathbf{a}^i)$ 
\Comment{Normalize and sort block-level scores}
% \vspace{0.1em}
\State $b_i \gets \min\!\left\{k \;\middle|\; \sum_{j=0}^k \mathbf{a}^i[j] \ge \gamma \right\} \,/\, |\mathbf{a}^i|$ 
\Comment{Determine budget ratio for selected head}
\State return $b_i$
\end{algorithmic}
\end{algorithm}

\vspace{-0.5em}
\section{Experiments}

\subsection{Experimental Setup}

\paragraph{Evaluation datasets.}
% Datasets
To thoroughly evaluate the performance of proposed method for long contexts,
we utilize two types of benchmarks: synthetic and real-world datasets.
% synthetic
For synthetic benchmarks,
we adopt the \textit{RULER} \citep{hsieh2024ruler},
which comprises a variety of synthetic tasks with token lengths ranging from 4K to 128K. 
% Why
This provides a direct way to evaluate the long-context capabilities of models
and the performance trade-off introduced by sparse attention.
% Real
Besides, We use \textit{InfiniteBench} \citep{zhang2024bench}
and \textit{LongBench-v2} \citep{bai2024longbench,bai2024longbench2}
as benchmarks for real-world tasks.
% Intro
Both include diverse long-text tasks,
such as QA and summarization,
providing a more comprehensive evaluation in practical applications.

\paragraph{Selected baselines.}
% Method Selected
To fairly evaluate the proposed \textit{ProxyAttn},
we compare it against the following mainstream block sparse attention methods as strong baselines:
(1) \textit{MInference} \citep{jiang2024minference}
enables dynamic sparse attention
by defining three unique patterns
% (A-shape, Vertical-Slash, and Block-Sparse)
and estimating the optimal indices for each pattern at inference time.
(2) \textit{FlexPrefill} \citep{lai2025flexprefill}
leverages input statistics to enable context-aware sparse pattern determination and Top-P block selection,
making the application of sparse attention more adaptable.
(3) \textit{XAttention} \citep{xu2025xattention}
mitigates the neglect of important tokens after pooling
by more effectively capturing the Vertical-Slash pattern
through the computation of the anti-diagonal.
(4) \textit{SeerAttention} \citep{gao2024seerattention}
employs a trainable MLP to process the latent information
from the pooled queries and keys,
which allows it to capture block importance more effectively than direct pooling.

% Configuration
For the hyperparameter settings of the baseline methods,
we strictly follow the optimal parameters reported in their original papers.
% details
Specifically,
MInference pre-configures patterns and budgets for each head.
% FlexPrefill
To ensure fair sparsity, we configure FlexPrefill 
with a $\gamma$ of 0.95 and a minimum budget of 2048.
% XAttention
While XAttention employs head budgets determined via offline search,
for models like Qwen where these budgets are not provided,
we directly set a threshold of 0.95 to ensure the method's performance.
% SeerAttention
SeerAttention implements dynamic block sparsity based on a threshold,
which we set to 5e-4 as original paper.

% RULER Benchmark
\begin{table}[t]
   \scriptsize
   % \small
   \centering
   \caption{Experimental results on RULER.
   \textbf{Bolded} values represent the best result under the same setting,
   while \underline{underlined} values indicate the second best.
   $*$ denotes methods that require additional training.
   Sparsity is calculated by weighting the ratio of different lengths by their token count.
   % A higher value indicates greater computational sparsity and higher efficiency.
   }
   \label{tab:ruler}
   \begin{tabularx}{\textwidth}{l*{10}{>{\centering\arraybackslash}X}}
   % \begin{tabular}{l*{10}c}
      \toprule
      \textbf{Method} & \textbf{Sparsity$\uparrow$}& \textbf{4k} & \textbf{8k} & \textbf{16k} & \textbf{32k} & \textbf{64k} & \textbf{128k} &
      \textbf{Avg.} & \textbf{wAvg.} \\
      \midrule
      \rowcolor{gray!10}
      \multicolumn{11}{c}{\textit{Llama3.1-8B-Instruct}} \\
      \midrule
      % \multirow{5}{*}{\rotatebox{90}{\scriptsize \textbf{Llama3.1-8B-Instruct}}} &
      FullAttention & 0.00 & 96.17 & 94.21 & 93.72 & 89.56 & 86.13 & 76.95 & 89.46 & 86.49\\
      \hdashline
      Minference & 0.70 & \textbf{96.23} & 94.22 & 93.92 & 88.82 & 84.64 & 70.74 & 88.09 & 84.25\\
      FlexPrefill & 0.72 & 96.03 & 94.07 & \textbf{94.35} & \textbf{92.05} & 85.21 & 75.12 & \underline{89.47} & 86.28\\
      XAttention & 0.69 & 96.19 & 94.27 & 93.99 & 91.04 & 85.15 & 72.64 & 88.88 & 85.35\\
      SeerAttention$^*$ & \underline{0.77} & \underline{96.20} & \underline{94.44} & 93.80 & 90.01 & 84.99 & 74.34 & 88.96 & 85.60\\
      \hdashline
      \rowcolor{blue!10}
      ProxyAttn ($\gamma$=0.90) & \textbf{0.80} & 95.76 & 94.32 & 93.52 & 91.05 & \underline{86.14} & \underline{75.75} & 89.42 & \underline{86.31}\\
      \rowcolor{blue!10}
      ProxyAttn ($\gamma$=0.95) & 0.69 & 96.07 & \textbf{94.57} & \underline{94.01} & \underline{91.55} & \textbf{86.63} & \textbf{78.25} & \textbf{90.18} & \textbf{87.43}\\
      \midrule
      % \multirow{5}{*}{\rotatebox{90}{\scriptsize \textbf{Qwen2.5-Ins-1M}}} &
      \rowcolor{gray!10}
      \multicolumn{11}{c}{\textit{Qwen2.5-7B-Instruct-1M}} \\
      \midrule
      FullAttention & 0.00 & 94.62 & 91.45 & 89.49 & 88.46 & 84.58 & 78.90 & 87.92 & 85.53\\
      \hdashline
      Minference & 0.63 & 94.41 & \underline{91.43} & 89.18 & \underline{87.56} & 82.73 & 76.60 & 86.99 & 84.20\\
      FlexPrefill & \underline{0.69} & 94.09 & 90.97 & \underline{89.27} & \textbf{87.59} & 81.85 & 76.24 & 86.67 & 83.85\\
      XAttention & 0.61 & 92.92 & 88.22 & 84.45 & 84.09 & 79.99 & 73.74 & 83.90 & 81.02\\
      \hdashline
      \rowcolor{blue!10}
      ProxyAttn ($\gamma$=0.90) & \textbf{0.74} & \underline{94.57} & 91.34 & 89.19 & 86.55 & \textbf{83.76} & \underline{76.80} & \underline{87.04} & \underline{84.32}\\
      \rowcolor{blue!10}
      ProxyAttn ($\gamma$=0.95) & 0.61 & \textbf{94.61} & \textbf{92.07} & \textbf{89.75} & 86.68 & \underline{83.57} & \textbf{77.09} & \textbf{87.30} & \textbf{84.53}\\
      \bottomrule
   % \end{tabular}
   \end{tabularx}
   \vspace{-1em}
\end{table}

\paragraph{Implementation details.}
% only prefill
Consistent with previous studies \cite{jiang2024minference,lai2025flexprefill},
we choose to evaluate sparse attention methods exclusively during the \textbf{pre-filling} stage,
which is the primary computational bottleneck,
while using full attention for the decoding stage.
% Setting
All performance and speed experiments are conducted on
an NVIDIA-H800-80GB platform.
% Kernel Computation
We fuse dot-product and max-pooling operators to reduce latency during the block estimation phase, following the kernel design of \citep{gao2024seerattention}.
% Block Sparse Attention
For the subsequent block sparse attention stage, we leverage the efficient implementation proposed by \citet{guo2024blocksparse}.
% Inference Speed Hardware

For the hyperparameter selection of our proposed ProxyAttn,
we adopt two settings with $\gamma$ values of 0.95 and 0.90,
respectively, while maintaining a stride of 4 and a minimum budget of 2048 tokens.
% Hyperparameter Experiments
See Appendix \ref{app:hyper} for more on hyperparameter experiments.
% Group
When choosing the number of proxy heads for estimation,
we selected 1 and 4 for the LLaMA and Qwen models,
considering the similarity of their attention heads.
% Llama
This indicates that a calculation using just one head 
accurately represent the behavior of all 32 heads
in the LLaMA3.1-8B model.
% More
We will discuss this further in Section \ref{ana:head}.

\subsection{Accuracy Results}

\paragraph{Results on synthetic tasks.}
Following the settings of \citet{hsieh2024ruler},
we also report weighted average scores (wAvg.),
which assigns linear weights to the results based on actual token lengths.
% Overall
The results for synthesis tasks with various length are presented in Table \ref{tab:ruler},
which shows that the proposed method has significant advantages over existing methods 
in terms of both sparsity and performance.
% Models with differ head
By sharing scores between attention heads,
ProxyAttn achieves excellent results on models with varying GQA settings,
including Llama3.1 and Qwen2.5.
% generalization
This highlights the strong generalization capabilities of the proposed method.

% Sparsity
Additionally, we evaluate the performance of the proposed method
at various sparsity rates by adjusting $\gamma$.
% Better
When ProxyAttn maintains a sparsity ratesimilar to other methods,
it achieves the best average performance on RULER,
even surpassing the performance of full attention on the Llama model.
% Faster
As the sparsity further increases,
the proposed method can still achieve comparable experimental results
while also reducing inference latency.
% Parameter
In contrast to the performance degradation XAttention faces when transferred to Qwen,
ProxyAttn can be quickly migrated to other models
without hyperparameter searching
by using online budget estimation.

\paragraph{Results on real-world tasks.}
% Info
In addition to synthetic tasks,
we also evaluate the performance of existing methods on real-world tasks
using benchmarks such as InfiniteBench \cite{zhang2024bench} and LongBench-v2 \cite{bai2024longbench2}.
% Overall still better than baseline.
As shown in Table \ref{tab:long}, despite the lower discriminability of real-world tasks compared to synthetic ones,
the proposed ProxyAttn still achieve the best average results
and even surpassed the performance of full attention.
% betterthen baseline
This demonstrates the potential of sparse attention methods for handling long documents.
% More
Full experimental results for the two datasets are provided in Appendix \ref{app:detailed_results}.

% InfiniteBench + LongBench-v2
\begin{table*}[t]
   \scriptsize
   \centering
   \caption{Main Results on InfiniteBench and LongBench-v2.
   LongBench results are reported without Chain-of-Thought (CoT),
   with the ``w. CoT" results shown in parentheses.
   The overall score is obtained by averaging the final scores
   from the two benchmarks.
   }
   \label{tab:long}
   \vspace{-1em}
   \begin{tabularx}{\textwidth}{l*{5}{>{\centering\arraybackslash}c}*{3}{>{\centering\arraybackslash}X}c}
   \toprule
   \multirow{2}{*}{\textbf{Methods}} & \multicolumn{5}{c}{\textbf{InfiniteBench}} & \multicolumn{3}{c}{\textbf{LongBench-v2 (w. CoT)}} & \multirow{2}{*}{\textbf{Overall}} \\
   \cmidrule(lr){2-6} \cmidrule(lr){7-9}
   & \textbf{En} & \textbf{Zh} & \textbf{Code} & \textbf{Math} & \textbf{Retri.} & \textbf{Short} & \textbf{Medium} & \textbf{Long} & \\
   \midrule
   \rowcolor{gray!10}
   \multicolumn{10}{c}{\textit{Llama3.1-8B-Instruct}} \\
   \midrule
   FullAttention & 32.09 & 13.65 & 23.35 & 33.43 & 84.84 & 33.90 (36.10) & 26.50 (30.70) & 25.00 (27.80) & 35.38 \\
   \hdashline
   Minference & 29.00 & 12.57 & \textbf{26.40} & \textbf{35.71} & 73.59 & \textbf{35.00} (\underline{38.90}) & 26.00 (\textbf{31.20}) & \textbf{30.60} (26.90) & 34.78 \\
   FlexPrefill & 30.75 & \textbf{13.69} & 23.60 & 32.29 & \underline{80.77} & 30.60 (36.70) & \textbf{27.90} (25.10) & 25.90 (\underline{28.70}) & 33.96 \\
   XAttention & \underline{31.45} & 13.28 & 22.84 & \underline{32.57} & 80.03 & \textbf{35.00} (35.00) & 26.00 (\textbf{31.20}) & 26.90 (27.80) & \underline{34.89} \\
   \rowcolor{blue!10}
   ProxyAttn & \textbf{33.21} & \underline{13.35} & \underline{25.13} & 32.00 & \textbf{81.49} & \underline{34.40} (\textbf{41.10}) & \underline{27.40} (\underline{27.90}) & \underline{29.60} (\textbf{29.60}) & \textbf{36.06} \\
   \midrule
   \rowcolor{gray!10}
   \multicolumn{10}{c}{\textit{Qwen2.5-7B-Instruct-1M}} \\
   \midrule
   FullAttention & 30.57 & 15.65 & 31.98 & 41.14 & 77.87 & 39.40 (38.90) & 28.40 (36.30) & 34.30 (33.30) & 38.22 \\
   \hdashline
   Minference & \underline{30.12} & 15.16 & 30.20 & \textbf{47.71} & \textbf{78.53} & 40.60 (\textbf{43.90}) & 26.50 (34.40) & \underline{28.70} (29.60) & \underline{37.91} \\
   FlexPrefill & \textbf{31.42} & \textbf{15.66} & 30.96 & 42.29 & 75.07 & \underline{41.70} (35.60) & \textbf{30.20} (\underline{34.90}) & 27.80 (\underline{31.50}) & 37.39 \\
   XAttention & 29.63 & \underline{15.56} & \underline{32.74} & \underline{43.71} & 70.40 & \underline{41.70} (\underline{42.20}) & 27.40 (33.50) & \textbf{32.40} (30.60) & 37.26 \\
   \rowcolor{blue!10}
   ProxyAttn & 29.75 & 15.36 & \textbf{32.99} & 41.71 & \underline{77.60} & \textbf{43.90} (40.00) & \underline{28.40} (\textbf{36.70}) & 27.80 (\textbf{32.40}) & \textbf{38.33} \\
   \bottomrule
   \end{tabularx}
   \vspace{-1em}
\end{table*}

\begin{figure}[t]
   \centering
   \includegraphics[width=\textwidth]{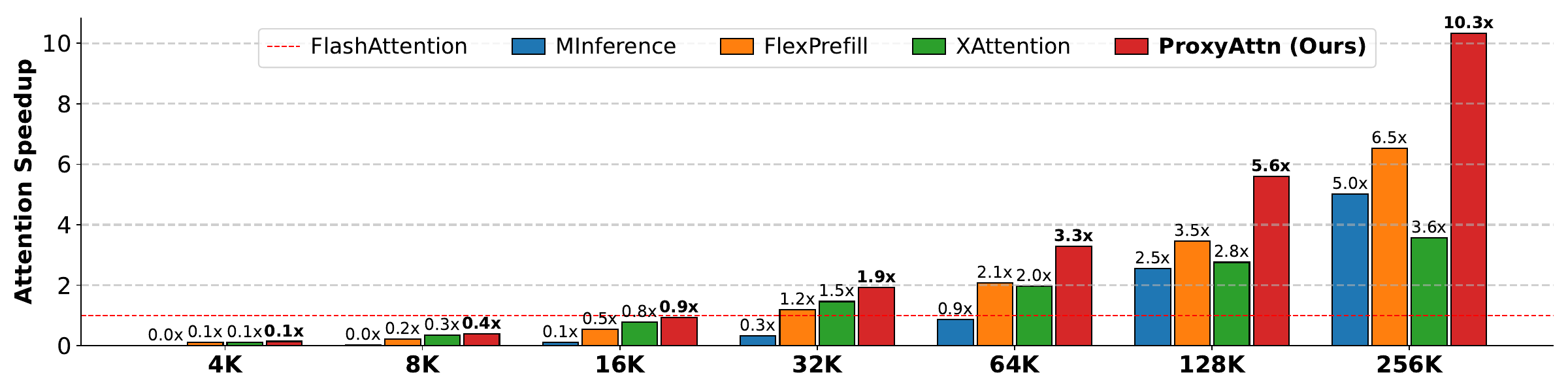}
   \caption{Kernel-level speedup of existing sparse attention methods with varying input lengths.}
   \label{fig:speedup}
   \vspace{-1em}
\end{figure}

\subsection{Efficiency Results}

\paragraph{Attention speedup.}
% Intro
To intuitively demonstrate the efficiency of our proposed method,
we first evaluate its attention acceleration ratio compared to other methods
at kernel level.
% Setting
Considering that the inductive bias of sparse attention is not effective with random vectors,
we cache the queries, keys, and values with a true distribution
from the RULER dataset for speed measurement.
% Results
The final speedup is obtained by averaging multiple runs across different layers.
We use the official implementation of FlashAttention \citep{dao2022flashattention,dao2023flashattention2}
as our baseline for testing,
and the experimental results are shown in Figure \ref{fig:speedup}.

As shown in the figure, sparse attention begins to show its speedup effect
when the input text exceeds 32K tokens.
% Results
Thanks to its fine-grained and low-cost block importance estimation,
ProxyAttn can achieve a lower sparsity (see Appendix \ref{app:sparse_ratio})
while maintaining performance,
thereby leading to a more significant speedup.
% Conclusion
It is worth noting that the proposed method can achieve a 10.3x attention acceleration with 256K tokens,
which would significantly improve the pre-filling speed of the model.

\paragraph{End-to-end prefilling speedup.}
The pre-fill latency of LLMs is also influenced by the MLP modules,
which diminishes the acceleration effect gained from optimizing the attention operation alone.
% Figure
As shown in Figure \ref{fig:intro},
we report the performance and Time to First Token (TTFT)
speedup of our proposed method under the RULER 128K setting.
% More
By adjusting the settings, the proposed ProxyAttn achieves a Pareto frontier of performance and efficiency
with its fine-grained block importance estimation.
% Act
With a minimal loss of model performance,
an end-to-end speedup of up to 2.4x is achieved.

\section{Analysis}

\begin{figure}[t]
   \centering
   \subcaptionbox{\label{fig:anz:group}}
   {\includegraphics[width=0.31\textwidth]{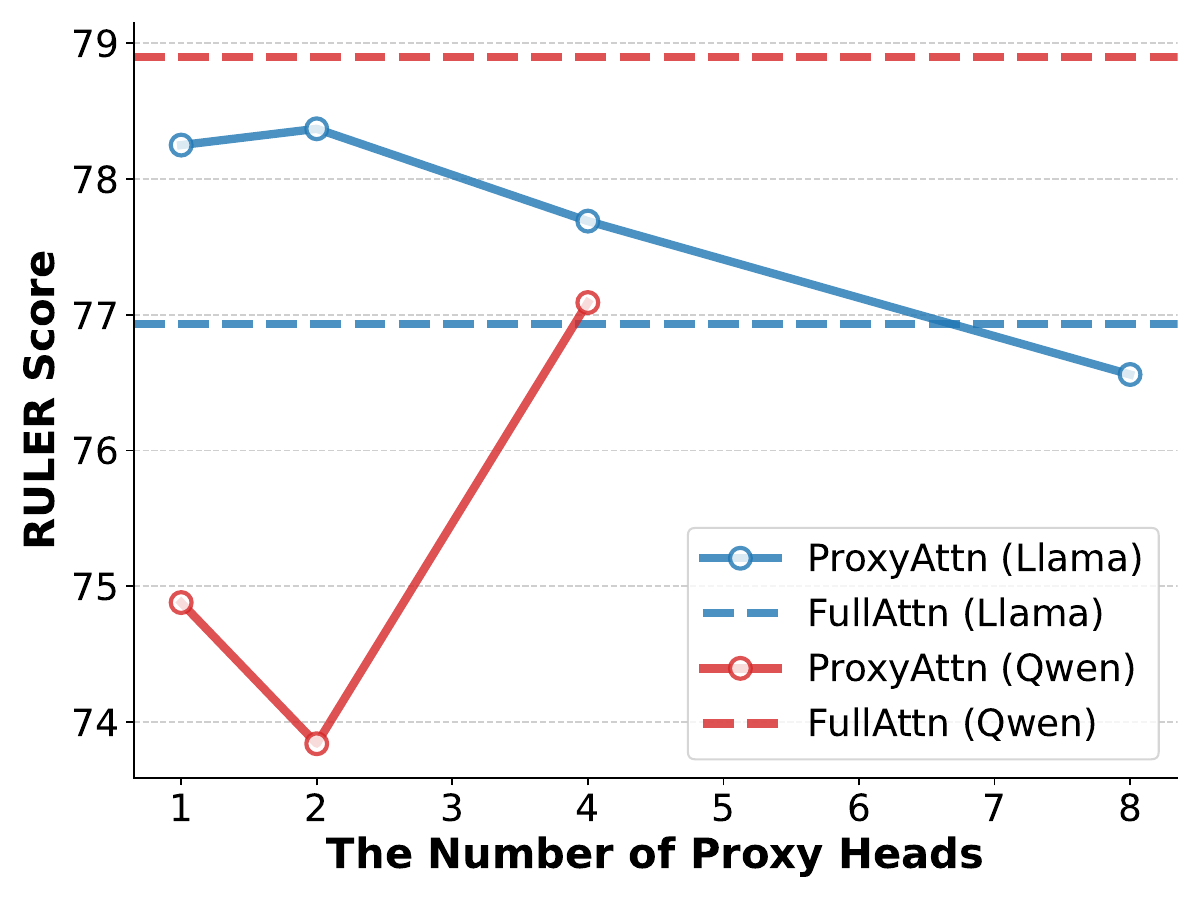}}
   \hfill
   \subcaptionbox{\label{fig:anz:esti}}
   {\includegraphics[width=0.31\textwidth]{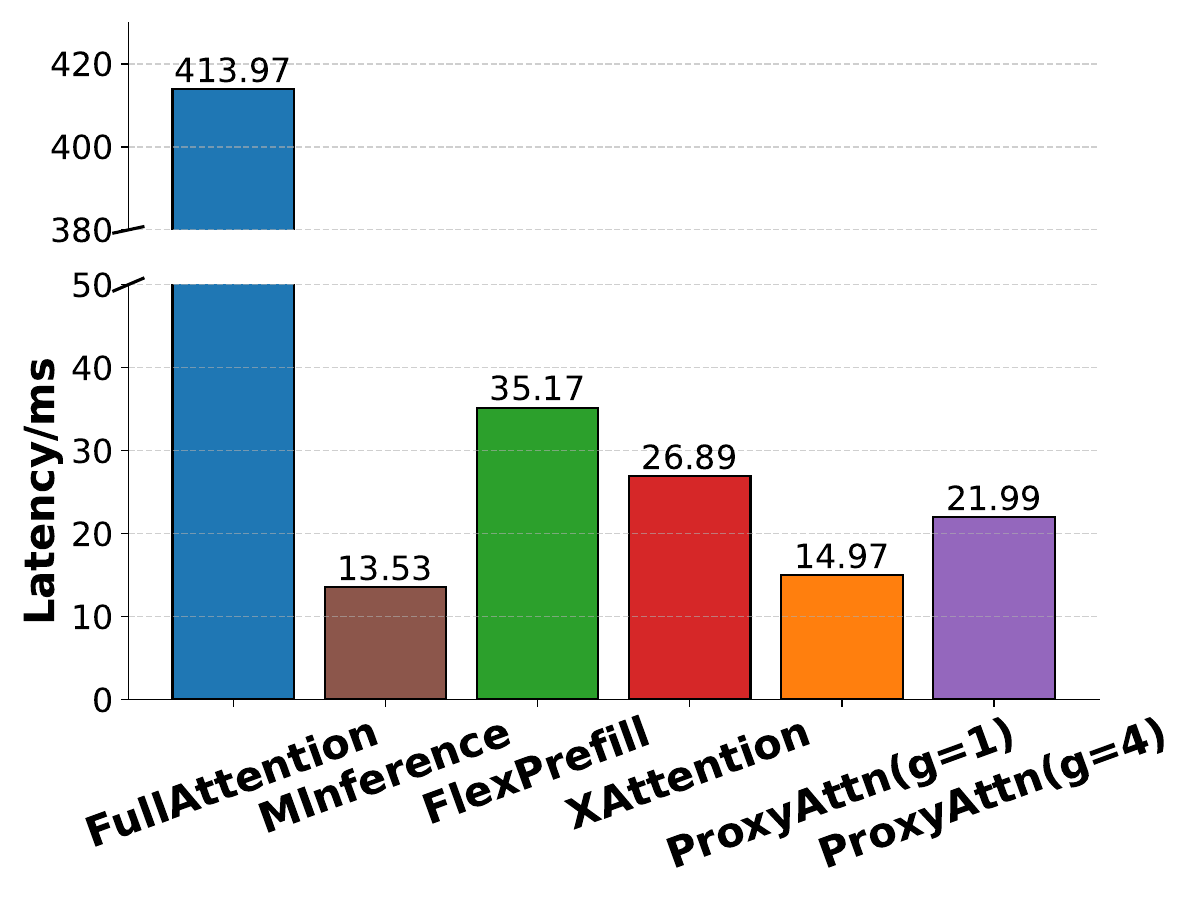}}
   \hfill
   \subcaptionbox{\label{fig:anz:sparse}}
   {\includegraphics[width=0.31\textwidth]{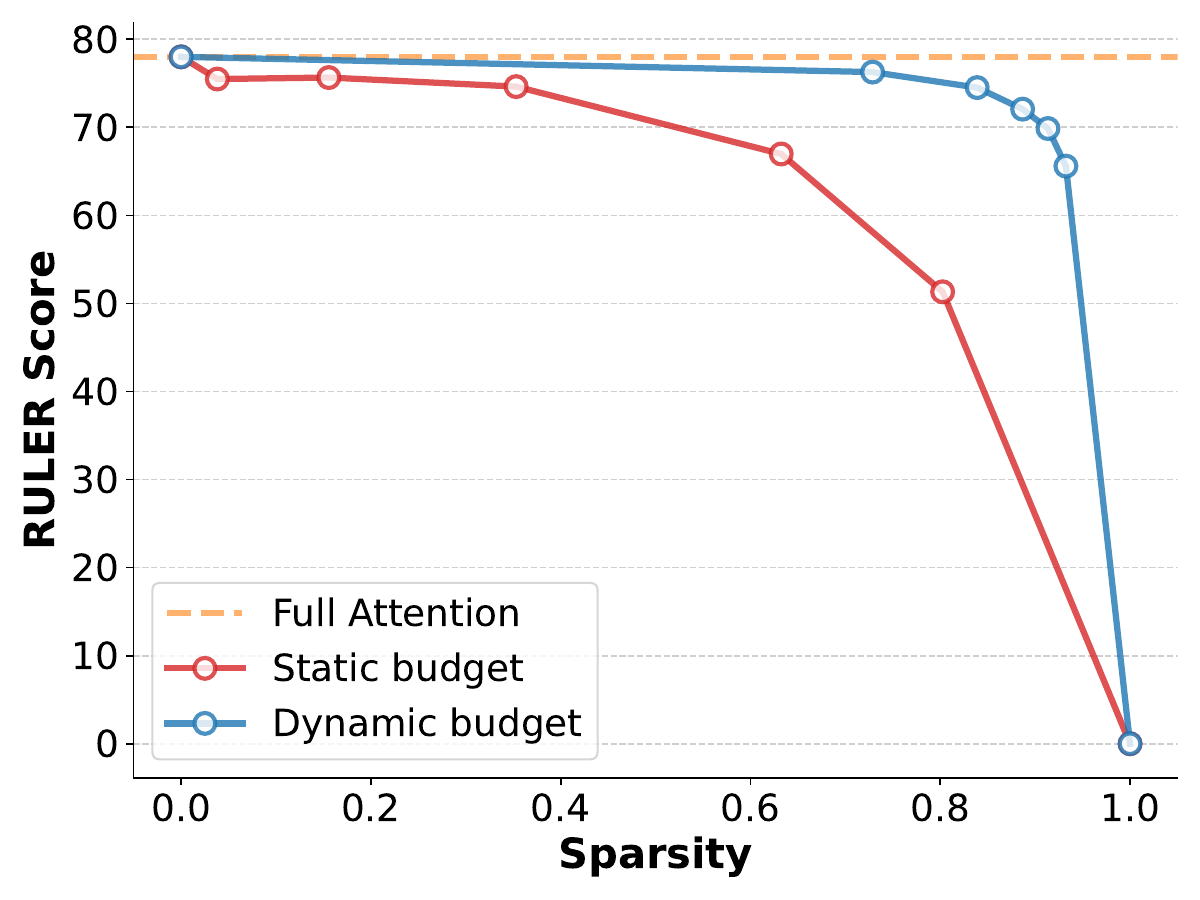}}
   \caption{Experimental analysis of proposed method.
(a) The performance of different numbers of proxy attention heads across various models.
(b) Latency for estimating block importance with 128K inputs using different methods.
The latency of all methods is less than 10\% of the Full Attention.
(c) Performance degradation with increasing sparsity rate
for different budget allocation methods.
}
\vspace{-1.5em}
   \label{fig:anz}
\end{figure}

\subsection{Attention Head Similarities Across Models}
\label{ana:head}
% Intro
The effectiveness of ProxyAttn is grounded in the existence of similarities among different heads.
% Why
To compare how this phenomenon varies across different models,
we conduct experiments on RULER 128K with a varying number of proxy heads.
% Figure
Figure \ref{fig:anz:group} shows the results
where we test models with different GQA configurations
by varying the number of proxy heads from 1 to the number of key heads.

% less is better
The multiple attention heads of the Llama model exhibit highly consistent performance.
% Llama
Consequently, a single proxy head can effectively represent all 32 heads, achieving performance that surpasses the baseline.
% Qwen
Furthermore, for the Qwen model, approximately 4 proxy heads are typically required to achieve satisfactory performance.
% Why
We hypothesize that this may be related to its very high GQA grouping ratio (seven queries per key).
% Now
Although increasing the number of proxy heads raises the latency of importance estimation,
the overall latency is not significant due to the use of strided dropout on the sequence dimension.

\begin{figure}[t]
   \centering
   \subcaptionbox{Conventionally computed attention scores.
   \label{fig:sink:before}}
   {\includegraphics[width=0.31\textwidth]{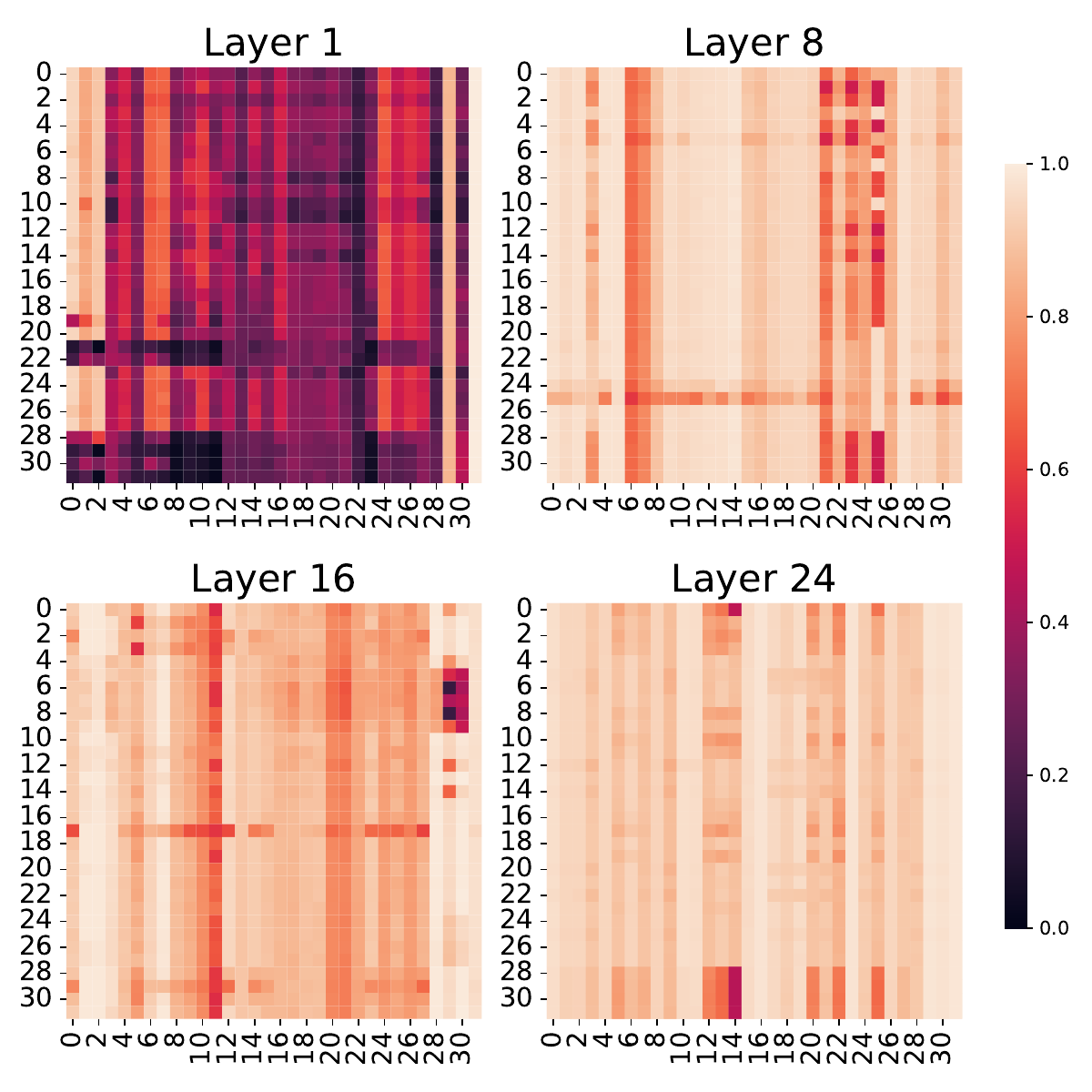}}
   \hfill
   \subcaptionbox{Attention scores after removing sink tokens (n=4).
   \label{fig:sink:after}}
   {\includegraphics[width=0.31\textwidth]{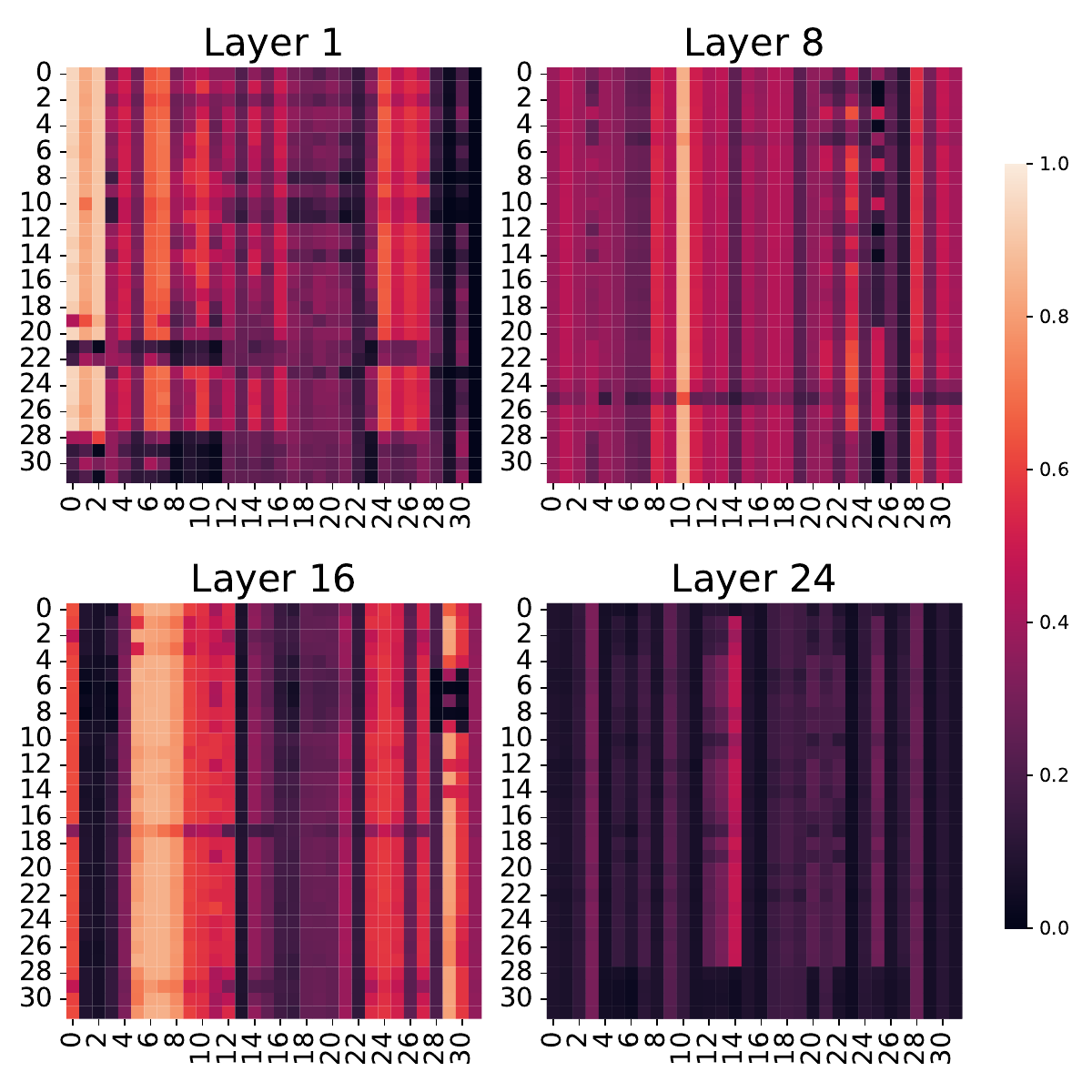}}
   \hfill
   \subcaptionbox{Normalized scores after removing sink tokens (n=4).
   \label{fig:sink:normalize}}
   {\includegraphics[width=0.31\textwidth]{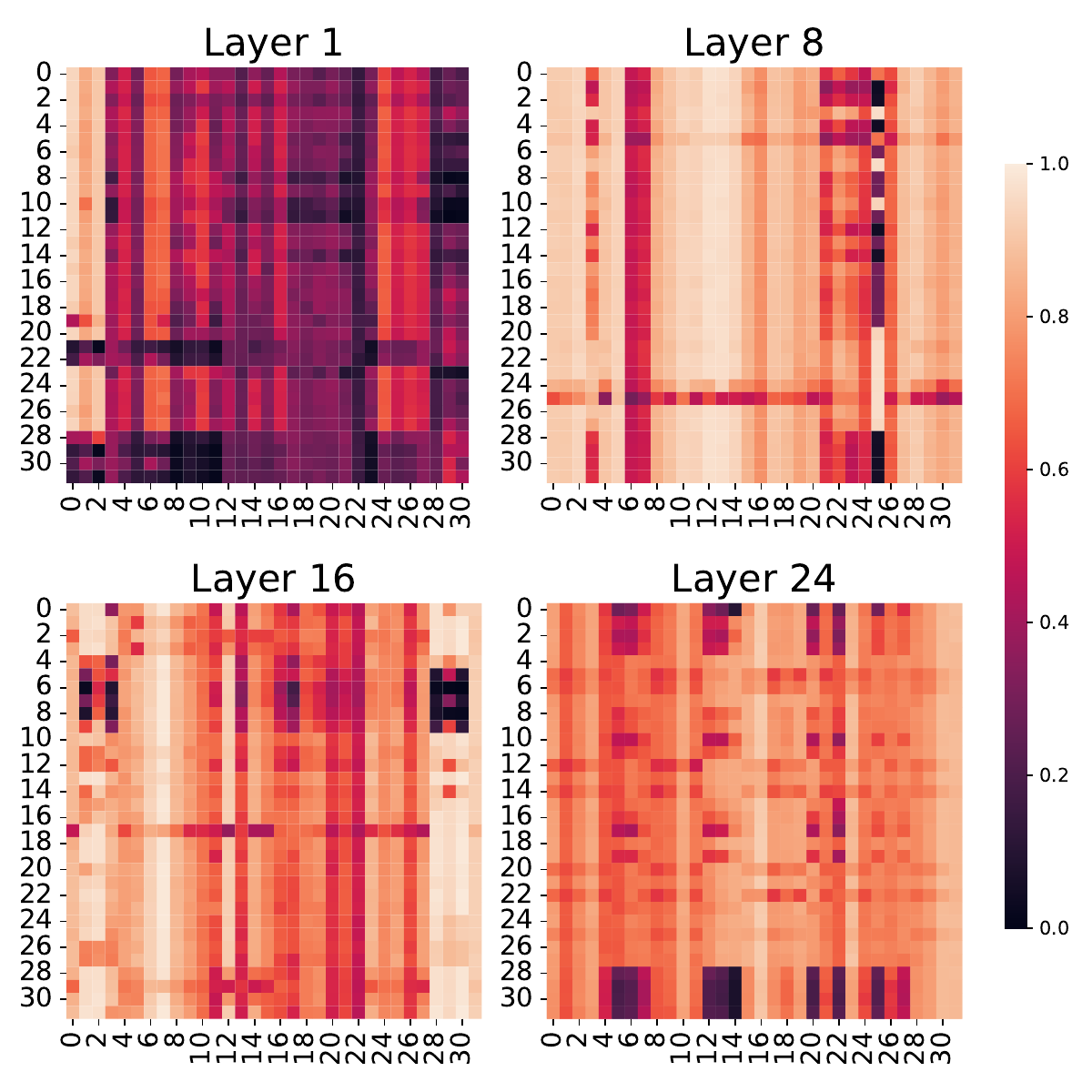}}
   \caption{
      Illustration of token overlap across different heads.
   }
   \label{fig:sink}
\end{figure}

\subsection{Comparison of Estimation Latency}
% Intro
Beyond achieving a higher sparsity rate,
the inherent latency of the block estimation algorithm
must also be factored into the overall algorithmic performance.
% What we do
To this end, we evaluate the overhead of existing sparse attention methods at the block estimation stage.
% Figure
As shown in Figure \ref{fig:anz:esti},
all methods achieve an estimation latency of less than 10\% of that of full attention.
% token
In particular, the proposed ProxyAttn reduces computational costs
by compressing the attention head dimension,
thereby allowing it to account for more fine-grained, token-level dot-products
while maintaining a latency comparable to other methods.
% Group
Furthermore, due to the use of dropout with a small stride,
increasing the number of proxy heads does not significantly increase the estimation latency,
which further enhances the generalizability of the method.

\subsection{The Effect of Dynamic Budget Allocation}
As discussed in \S \ref{sec:finding},
the sparsity among attention heads is highly variable.
% Static
Consequently, if a static Top-K allocation method is employed,
it can lead to excessive information loss in some dense heads,
which in turn impacts the final performance.
% As Figure
To validate our hypothesis, we compare the performance degradation of static and dynamic methods
as the sparsity rate increased, using 128K-token inputs.
% Results
As shown in Figure \ref{fig:anz:sparse},
the proposed method significantly outperforms the static baseline,
exhibiting a notable performance degradation only when the sparsity exceeds 90\%.
% Conclusion
This highlights the importance of independently estimating the budget for each head.

\vspace{-0.5em}
\subsection{Model Scaling Experiments}
\begin{wraptable}{r}{0.45\textwidth}
   \small
   \centering
   \vspace{-1.5em}
   \caption{Ruler results on Llama3.1-70B-Instruct
   with 64 attention heads.}
   \label{tab:scaling}
   \begin{tabular}{ccc}
      \toprule
      \textbf{Method} & \textbf{RULER} & \textbf{TTFT/s} \\
      \midrule
      FullAttention & 65.03 & 91.84 (1.00$\times$)\\
      \hdashline
      MInference & 60.33 & 82.90 (1.11$\times$)\\
      XAttention & 58.59 & 60.57 (1.52$\times$)\\
      ProxyAttn & \textbf{62.23} & \textbf{44.23 (2.08$\times$)}\\
      \bottomrule
   \end{tabular}
   \vspace{-1em}
\end{wraptable}
Given that larger models typically have more attention heads,
we scale our experiments to a 70B model to validate the effectiveness of our method with a greater number of attention heads.
% Compare
To ensure a fair comparison,
we also conducte experiments on several baselines that are compatible with the 70B model,
while maintaining the same configurations as primary experiments.
% Reults
As shown in Table \ref{tab:scaling}, even with the number of attention heads increased to 64, we can still achieve effective block importance estimation using a single proxy head.
% Report
While maintaining optimal performance,
we can achieve a 2.08x end-to-end prefilling speedup for Llama3.1-70B on 128K sequences.

\subsection{Effect of Sink Tokens}
% Motivation
Recent studies \citep{xiao2023efficient} have found that sink tokens,
typically located at the beginning of a sequence,
often receive a large amount of attention score.
% Sink
To eliminate the influence of sink tokens,
we conducte a revised observational experiment based on \S \ref{sec:finding},
removing attention scores to the sink tokens.
% For
For a given token, let its attention scores be $\mathcal{A}$,
with shape $\left(\texttt{head\_num}, \texttt{seq\_len}\right)$.
The overlap scores at $\left(\texttt{i}, \texttt{j}\right)$ in the figure are computed as follows:

\begin{equation}
\begin{aligned}
\text{token\_index} &= \text{topk}(\mathcal{A}[i], k=1024).\text{indices} \\
\text{Score}[i, j] &= \text{sum}(\mathcal{A}[j, \text{token\_index}])
\end{aligned}
\end{equation}

% conclusion
The experimental results are shown in Figure \ref{fig:sink}.
% Sink Tokens' score grows
It can be observed that as the layer depth increases, the attention scores received by sink tokens grow progressively larger.
% Consistency
After re-normalizing the attention,
there is still substantial overlap across different heads.
This indicates that the improvements provided by the proxy head are not merely a result of shared sink tokens,
but also reflect its capacity to act as a proxy at the global level.

\section{Related Work}
% \subsection{Sparse Attention}
\paragraph{Multi-head attention.}
% info
While offering powerful modeling capabilities,
Multi-Head Attention (MHA) \citep{vaswani2017attention} 
inherently suffers from efficiency issues.
% MHA & GQA & MHA & MLA
Many recent studies \citep{zheng2024attention} aim to refine the attention mechanism
from the perspective of multiple heads.
% MHA & GQA
Multi-Query Attention (MQA) \citep{shazeer2019fast} and Grouped-Query Attention (GQA) \citep{ainslie2023gqa}
significantly reduce the memory footprint of the KV cache
by having multiple queries share a single set of keys and values.
% MoH
Multi-head Latent Attention (MLA) \citep{liu2024deepseek}
implicitly groups via low-rank projection,
achieving a further reduction in KV Cache memory consumption.

In addition, other work investigates the differential performance
across various heads.
MoA \citep{fu2024moa} achieves efficient attention computation by assigning diverse,
sparse attention patterns to each head in every layer.
DuoAttention \citep{xiao2024duoattention}
utilizes an optimization method to effectively identify sliding-window heads
and subsequently performs the corresponding sparse computation.
% Conclusion
Beyond the sparsity differences discussed above,
the findings of this paper also indicate a potential consistency across multiple heads,
which can be leveraged for designing subsequent efficient attention architectures.

\paragraph{Sparse attention.}
% Training-free
Exploiting the inherent sparsity of the attention mechanism,
MInference \citep{jiang2024minference} and FlexPrefill \citep{lai2025flexprefill}
achieve efficient block-sparse attention by computing a dynamic sparse mask.
% SeerAttention
SeerAttention \citep{gao2024seerattention} introduces a trainable MLP to extract information from different blocks,
leading to a more accurate estimation.
% Fine-grained
XAttention extends the pooling-based methods by incorporating a more fine-grained anti-diagonal scoring mechanism,
which leads to a better capture of the heuristic ``vertical and slash" patterns.
% SharePrefill
Similarly inspired by inter-head similarity, 
SharePrefill \citep{peng2025accelerating} performs offline clustering of attention heads.
% What
By fully sharing sparse patterns among similar heads,
it achieves more accurate block attention estimation.

% Training-based
However, directly applying sparse attention to LLMs still incurs an unavoidable performance degradation.
% Info
Some approaches \citep{team2025minicpm4} attempt to introduce optimization objectives during training
to induce native sparsity of LLMs.
NAS \citep{yuan2025native}
achieves a native sparse attention that surpasses full attention
by integrating three modes: compression, selection, and sliding window.
% MoBA
Referencing the principles of MoE (Mixture-of-Experts),
MOBA \citep{lu2025moba} achieves a mix of block sparse attention without introducing any additional parameters.

\section{Conclusion}
% Overall
In this paper, we propose ProxyAttn, a training-free method for sparse attention.
% Performance
Considering the similarity among multiple attention heads,
we leverage proxy heads to efficiently estimate the attention scores for all heads.
% Budget
When integrated with dynamic budget allocation,
ProxyAttn outperforms existing approaches across multiple datasets and models.
% Future Work
We will further investigate the utility of the proxy head specifically within the decoding phase of LLMs.
% with the ultimate goal of enhancing the efficiency of the Attention mechanism.

\section*{Acknowledgements}

We gratefully acknowledge the support of the National Natural Science Foundation of China (NSFC) via grant 62236004 and 62476073.

\bibliography{iclr2026_conference}
\bibliographystyle{iclr2026_conference}

\appendix
\section{AI Tools}
In this manuscript, AI tools such as ChatGPT were used solely for refining the text
and did not contribute to the content or ideas presented.

\section{Hyper-parameters}
\label{app:hyper}
We present the hyperparameter search experiments related to ProxyAttn in this section.
% final
Ultimately, we chose a stride of 4 and a minimum budget of 0 for LLaMA and 2048 for Qwen
for our main experiments.

\subsection{Effect of Gamma and Min Budget}
\begin{table}[h]
   \scriptsize
   \centering
   \caption{The hyperparameter search experiments concerning sparsity on RULER Benchmark.}
   \begin{tabular}{lcccccccccc}
      \toprule
      \textbf{$\gamma$} & \textbf{min budget} & \textbf{sparsity} & \textbf{4k} & \textbf{8k} & \textbf{16k} & \textbf{32k} & \textbf{64k} & \textbf{128k} & \textbf{Avg.} & \textbf{wAvg.}\\
      \midrule
      \rowcolor{gray!10}
      \multicolumn{11}{c}{\textit{Llama3.1-8B-Instruct}} \\
      \midrule
      0.95 & 0 & 0.69 & 96.07 & 94.57 & 94.01 & 91.55 & 86.63 & 78.25 & \textbf{90.18} & \textbf{87.43} \\
      0.95 & 1024 &  0.69 & 95.75 & 94.04 & 93.65 & 90.66 & 86.05 & 76.92 & 89.51 & 86.63 \\
      0.95 & 2048 &  0.67 & 95.96 & 94.17 & 93.60 & 90.37 & 85.96 & 75.94 & 89.33 & 86.29\\
      0.90 & 0 & 0.81 & 95.79 & 94.30 & 93.15 & 90.91 & 85.62 & 76.16 & 89.32 & 86.31\\
      0.90 & 1024 & 0.80 & 95.82 & 94.23 & 93.31 & 90.18 & 84.72 & 73.63 & 88.65 & 85.25\\
      0.90 & 2048 & 0.77 & 95.96 & 94.25 & 93.19 & 90.14 & 84.64 & 73.09 & 88.55 & 85.06\\
      \midrule
      \rowcolor{gray!10}
      \multicolumn{11}{c}{\textit{Qwen2.5-7B-Instruct-1M}} \\
      \midrule
      0.95 & 0 & 0.63 & 93.91 & 91.26 & 90.16 & 86.89 & 83.46 & 76.99 & 87.11 & 84.46\\
      0.95 & 1024 & 0.62 & 94.22 & 90.91 & 90.10 & 87.25 & 83.43 & 77.51 & 87.24 & 84.65 \\
      0.95 & 2048 & 0.61 & 94.61 & 92.07 & 89.75 & 86.68 & 83.57 & 77.09 & \textbf{87.30} & \textbf{84.53} \\
      0.90 & 0 & 0.76 & 93.59 & 90.27 & 88.34 & 86.19 & 82.10 & 75.84 & 86.06 & 83.31\\
      0.90 & 1024 & 0.76 & 94.07 & 90.62 & 89.32 & 86.48 & 82.65 & 74.71 & 86.31 & 83.37\\
      0.90 & 2048 & 0.74 & 94.57 & 91.34 & 89.19 & 86.55 & 83.76 & 76.80 & 87.04 & 84.32\\
      \bottomrule
   \end{tabular}
\end{table}

\subsection{Effect of Stride}
\label{app:stride}
As shown in Table \ref{tab:stride},
a slight increase in the stride does not significantly impair performance
due to the local similarity of attention,
but it substantially reduces the estimation overhead.

\begin{table}[h!]
   \small
   \centering
   \caption{The impact of varying strides on the proposed method.}
   \label{tab:stride}
   \begin{tabular}{lcccccccccc}
      \toprule
      \textbf{stride} & \textbf{$\gamma$} & \textbf{min budget} & \textbf{4k} & \textbf{8k} & \textbf{16k} & \textbf{32k} & \textbf{64k} & \textbf{128k} & \textbf{Avg}\\
      \midrule
      \rowcolor{gray!10}
      \multicolumn{10}{c}{\textit{Llama3.1-8B-Instruct}} \\
      \midrule
      \multicolumn{3}{l}{Full Attention} & 95.63 & 92.27 & 91.60 & 87.66 & 84.85 & 76.17 & 88.03 \\
      \hdashline
      1 & \multirow{3}{*}{0.95} & \multirow{3}{*}{0} & 95.43 & 92.16 & 90.94 & 89.25 & 84.98 & 74.44 & 87.87 \\
      2& &  & 95.32 & 92.13 & 91.46 & 89.29 & 85.24 & 75.56 & 88.17 \\
      4& &  & 95.16 & 92.89 & 91.84 & 89.69 & 84.64 & 76.27 & 88.42\\
      \midrule
      \rowcolor{gray!10}
      \multicolumn{10}{c}{\textit{Qwen2.5-7B-Instruct-1M}} \\
      \midrule
      \multicolumn{3}{l}{Full Attention} & 93.35 & 89.94 & 87.81 & 87.56 & 85.33 & 76.89 & 86.81 \\
      \hdashline
      1 &\multirow{3}{*}{0.95} & \multirow{3}{*}{2048} & 93.67 & 89.35 & 88.23 & 87.06 & 85.41 & 76.24 & 86.66 \\
      2& &  & 93.67 & 89.88 & 88.54 & 87.66 & 85.13 & 75.10 & 86.66 \\
      4& &  & 93.57 & 90.48 & 88.30 & 86.55 & 84.21 & 74.97 & 86.35 \\
      \bottomrule
   \end{tabular}
\end{table}

\section{Detailed Results on Real-World Datasets}
\label{app:detailed_results}

Due to space limitations, the main text reports the average performance
across two datasets.
The complete experimental results are presented in this section.

\begin{table*}[h]
   \scriptsize
   \centering
   \caption{Main results on InfiniteBench across various models.}
   \label{tab:InfinBench}
   % \begin{tabularx}{\textwidth}{l*{11}{>{\centering\arraybackslash}X}}
   \begin{tabular}{l*{11}c}
      \toprule
      \multirow{2}{*}{\textbf{Methods}} & \multicolumn{4}{c}{\textbf{En}} & \multicolumn{1}{c}{\textbf{Zh}} & \multicolumn{1}{c}{\textbf{Code}} & \multicolumn{1}{c}{\textbf{Math}} & \multicolumn{3}{c}{\textbf{Retrieval}} & 
      \multirow{2}{*}{\textbf{Avg.}}\\
      \cmidrule(lr){2-5} \cmidrule(lr){6-6} \cmidrule(lr){7-7} \cmidrule(lr){8-8} \cmidrule(lr){9-11}
       & \textbf{Sum} & \textbf{QA} & \textbf{MC} & \textbf{Dia} & \textbf{QA} & \textbf{Debug} & \textbf{Find} & \textbf{PassKey} & \textbf{Number} & \textbf{KV}  & \\
      \midrule
      \rowcolor{gray!10}
      \multicolumn{12}{c}{\textit{Llama3.1-8B-Instruct}} \\
      \midrule
      FullAttention & 27.12 & 14.37 & 66.38 & 20.50 & 13.65 & 23.35 & 33.43 & 100.00 & 99.32 & 55.20 & 45.33 \\
      \hdashline
      Minference & 25.51 & 14.30 & 60.70 & 15.50 & 12.57 & 26.40 & 35.71 & 100.00 & 97.97 & 22.80 & 41.15 \\
      FlexPrefill & 26.54 & 14.15 & 66.81 & 15.50 & 13.69 & 23.60 & 32.29 & 100.00 & 99.32 & 43.00 & 43.49 \\
      XAttention & 26.87 & 14.00 & 69.43 & 15.50 & 13.28 & 22.84 & 32.57 & 100.00 & 99.49 & 40.60 & 43.46 \\
      \rowcolor{blue!10}
      ProxyAttn & 26.78 & 14.85 & 67.69 & 23.50 & 13.35 & 25.13 & 32.00 & 100.00 & 99.66 & 44.80 & \textbf{44.78} \\
      \midrule
      \rowcolor{gray!10}
      \multicolumn{12}{c}{\textit{Qwen2.5-7B-Instruct-1M}} \\
      \midrule
      FullAttention & 23.48 & 16.86 & 65.94 & 16.00 & 15.65 & 31.98 & 41.14 & 100.00 & 100.00 & 33.60 & 44.47 \\
      \hdashline
      Minference & 24.37 & 15.87 & 67.25 & 13.00 & 15.16 & 30.20 & 47.71 & 100.00 & 100.00 & 35.60 & \textbf{44.92}\\
      FlexPrefill & 23.68 & 15.50 & 65.50 & 21.00 & 15.66 & 30.96 & 42.29 & 100.00 & 100.00 & 25.20 & 43.98\\
      XAttention & 22.64 & 15.99 & 66.38 & 13.50 & 15.56 & 32.74 & 43.71 & 100.00 & 100.00 & 11.20 & 42.17\\
      \rowcolor{blue!10}
      ProxyAttn & 23.16 & 15.26 & 65.07 & 15.50 & 15.36 & 32.99 & 41.71 & 100.00 & 100.00 & 32.80 & \underline{44.19}\\
      \bottomrule
   \end{tabular}
   % \end{tabularx}
\end{table*}

\begin{table*}[h]
   \scriptsize
   % \small
   \centering
   \caption{Main results on LongBench-v2 across various models.}
   \label{tab:LongBench}
   % \begin{tabularx}{\textwidth}{l*{8}{>{\centering\arraybackslash}X}}
   \begin{tabular}{l*{8}{c}}
   \toprule
   \multirow{2}{*}{\textbf{Method}} & \multicolumn{2}{c}{\textbf{Short}} & \multicolumn{2}{c}{\textbf{Medium}} & \multicolumn{2}{c}{\textbf{Long}} & \multicolumn{2}{c}{\textbf{Overall}} \\
   \cmidrule(lr){2-3} \cmidrule(lr){4-5} \cmidrule(lr){6-7} \cmidrule(lr){8-9}
   & w.o. Cot &  w. Cot & w.o. Cot &  w. Cot & w.o. Cot &  w. Cot & w.o. Cot &  w. Cot \\ 
   \midrule
   \rowcolor{gray!10}
   \multicolumn{9}{c}{\textit{Llama3.1-8B-Instruct}} \\
   \midrule
   FullAttention & 33.90 & 36.10 & 26.50 & 30.70 & 25.00 & 27.80 & 28.80 & 32.00 \\
   \hdashline
   Minference & 35.00 & 38.90 & 26.00 & 31.20 & 30.60 & 26.90 & 30.20 & \textbf{33.00} \\
   FlexPrefill & 30.60 & 36.70 & 27.90 & 25.10 & 25.90 & 28.70 & 28.40 & 30.00 \\
   XAttention & 35.00 & 35.00 & 26.00 & 31.20 & 26.90 & 27.80 & 29.40 & 31.80 \\
   \rowcolor{blue!10}
   ProxyAttn & 34.40 & 41.10 & 27.40 & 27.90 & 29.60 & 29.60 & \textbf{30.40} & \textbf{33.00} \\
   \midrule
   \rowcolor{gray!10}
   \multicolumn{9}{c}{\textit{Qwen2.5-7B-Instruct-1M}} \\
   \midrule
   FullAttention & 39.40 & 38.90 & 28.40 & 36.30 & 34.30 & 33.30 & 33.60 & 36.60 \\
   \hdashline
   Minference & 40.60 & 43.90 & 26.50 & 34.40 & 28.70 & 29.60 & 32.00 & 36.80 \\
   FlexPrefill & 41.70 & 35.60 & 30.20 & 34.90 & 27.80 & 31.50 & \textbf{33.80} & 34.40 \\
   XAttention & 41.70 & 42.20 & 27.40 & 33.50 & 32.40 & 30.60 & 33.60 & 36.00 \\
   \rowcolor{blue!10}
   ProxyAttn & 43.90 & 40.00 & 28.40 & 36.70 & 27.80 & 32.40& \textbf{33.80} & \textbf{37.00} \\
   \bottomrule
   % \end{tabularx}
   \end{tabular}
\end{table*}

\section{Sparsity Rates}
\label{app:sparse_ratio}
The sparsity rates achieved by different methods using the optimal configuration on RULER.
A higher sparsity rate leads to a greater acceleration effect.

\begin{table}[h]
\small
\centering
\caption{Sparsity rates achieved by different sparse attention methods
across varying input lengths.}
\begin{tabular}{lcccccc}
\toprule
\textbf{Model} & \textbf{4K} & \textbf{8K} & \textbf{16K} & \textbf{32K} & \textbf{64K} & \textbf{128K} \\
\midrule
\rowcolor{gray!10}
\multicolumn{7}{c}{\textit{Llama3.1-8B-Instruct}} \\
\midrule
MInference      & 6.15  & 11.62 & 22.60 & 34.13 & 50.44 & 72.24 \\
FlexPrefill     & 19.76 & 43.62 & 57.82 & 65.15 & 71.86 & 75.10 \\
XAttention      & 29.10 & 39.81 & 49.91 & 58.96 & 68.42 & 73.20 \\
% ProxyAttn-0.95  & 36.59 & 48.41 & 58.00 & 63.86 & 71.04 & 72.82 \\
ProxyAttn & \textbf{49.63} & \textbf{63.21} & \textbf{73.31} & \textbf{78.27} & \textbf{83.19} & \textbf{83.86} \\
\midrule
\rowcolor{gray!10}
\multicolumn{7}{c}{\textit{Qwen2.5-7B-Instruct-1M}} \\
\midrule
MInference          & 25.99 & 32.02 & 39.54 & 50.35 & 61.82 & 72.44 \\
FlexPrefill         & 19.39 & \textbf{45.02} & 59.26 & 65.50 & 68.39 & 74.35 \\
XAttention          & \textbf{34.11} & 44.47 & 51.54 & 56.99 & 60.55 & 65.32 \\
% ProxyAttn-0.95-qwen & 11.86 & 34.95 & 49.94 & 57.23 & 60.31 & 67.02 \\
ProxyAttn & 13.48 & 42.43 & \textbf{60.78} & \textbf{70.23} & \textbf{74.12} & \textbf{79.52} \\
\bottomrule
\end{tabular}
\end{table}

\section{Observations in Both Synthetic and Real-World Tasks}
\begin{figure}[h]
   \centering
   \subcaptionbox{Layer 0
   \label{fig:fake:1}}
   {\includegraphics[width=0.24\textwidth]{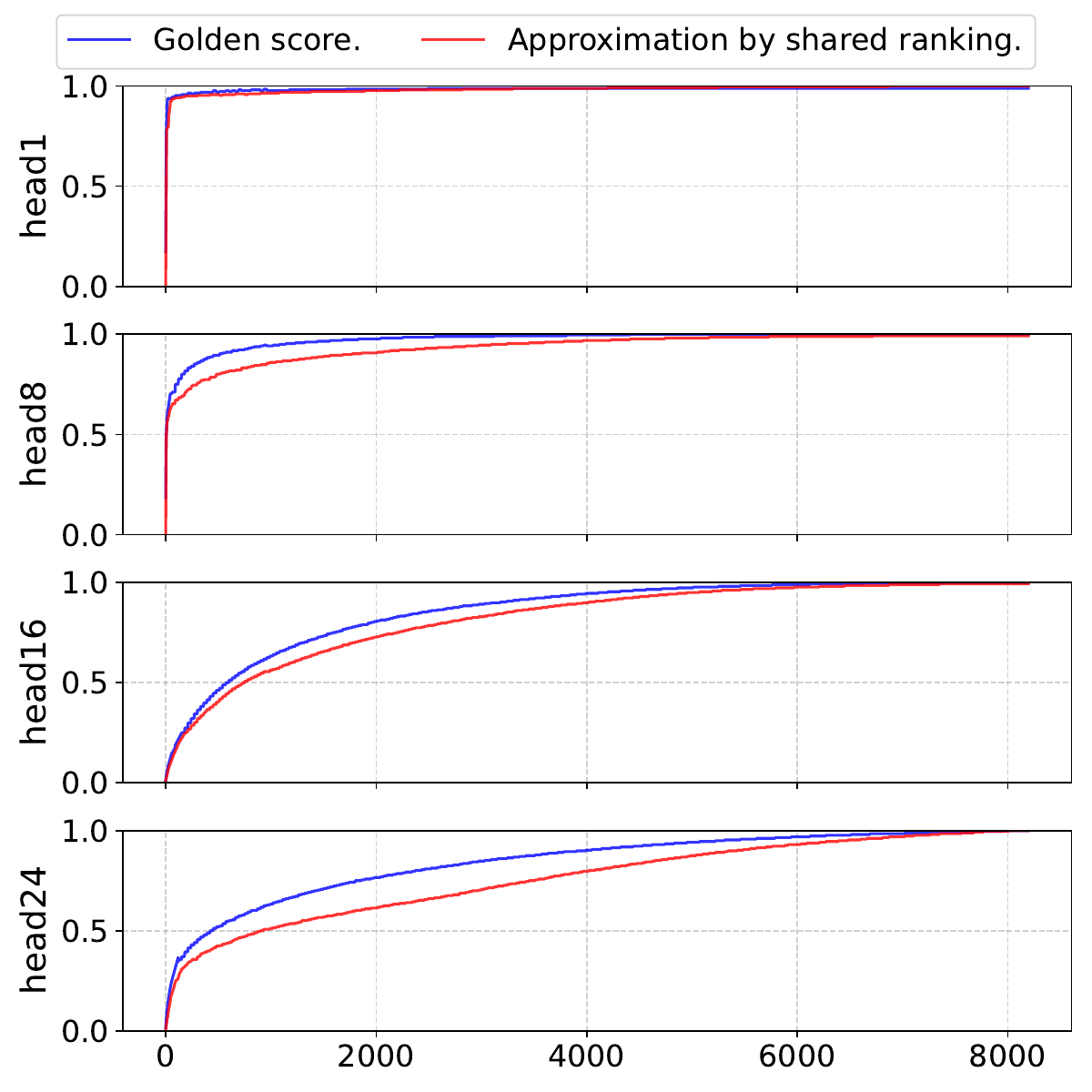}}
   \hfill
   \subcaptionbox{Layer 7
   \label{fig:fake:2}}
   {\includegraphics[width=0.24\textwidth]{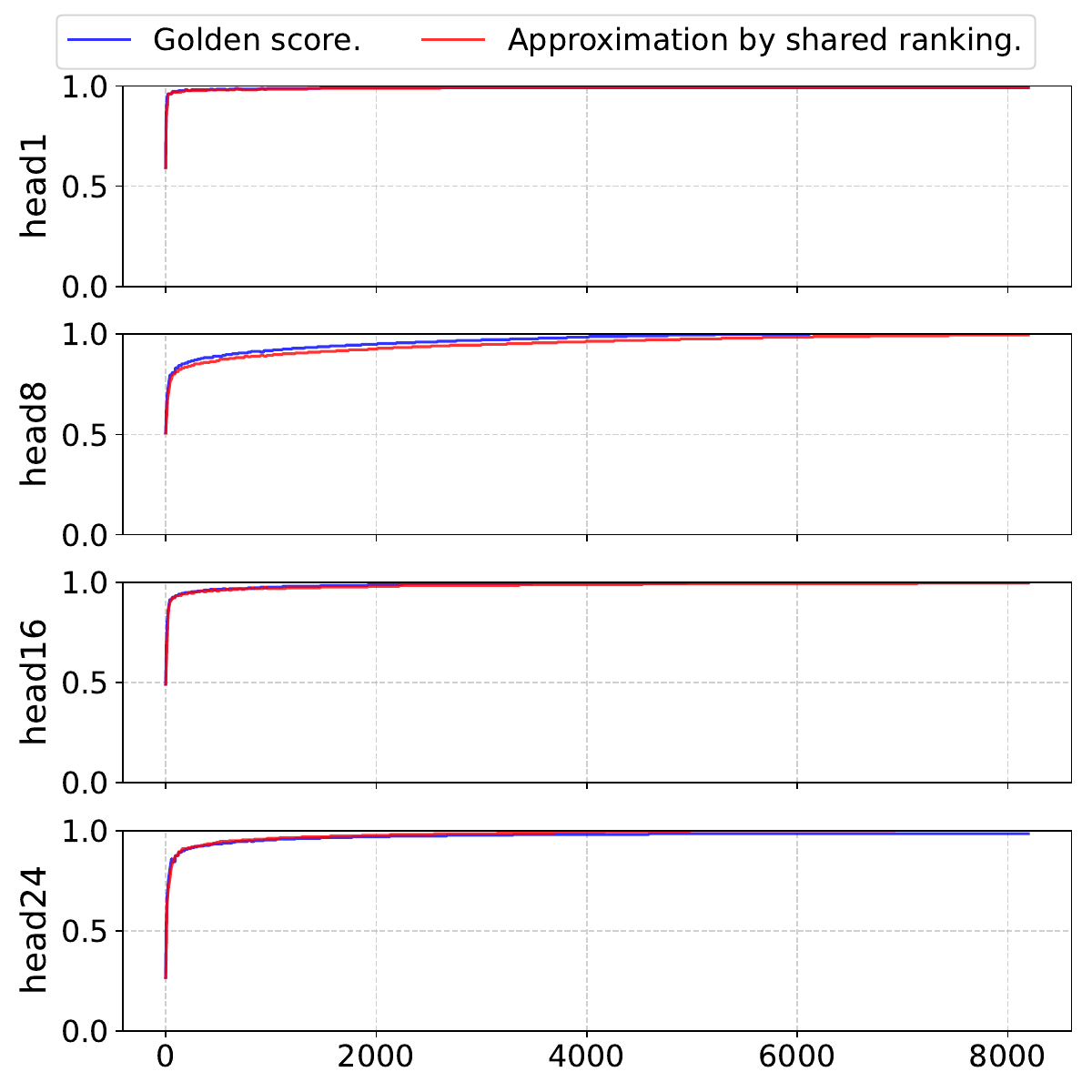}}
   \hfill
   \subcaptionbox{Layer 15
   \label{fig:fake:3}}
   {\includegraphics[width=0.24\textwidth]{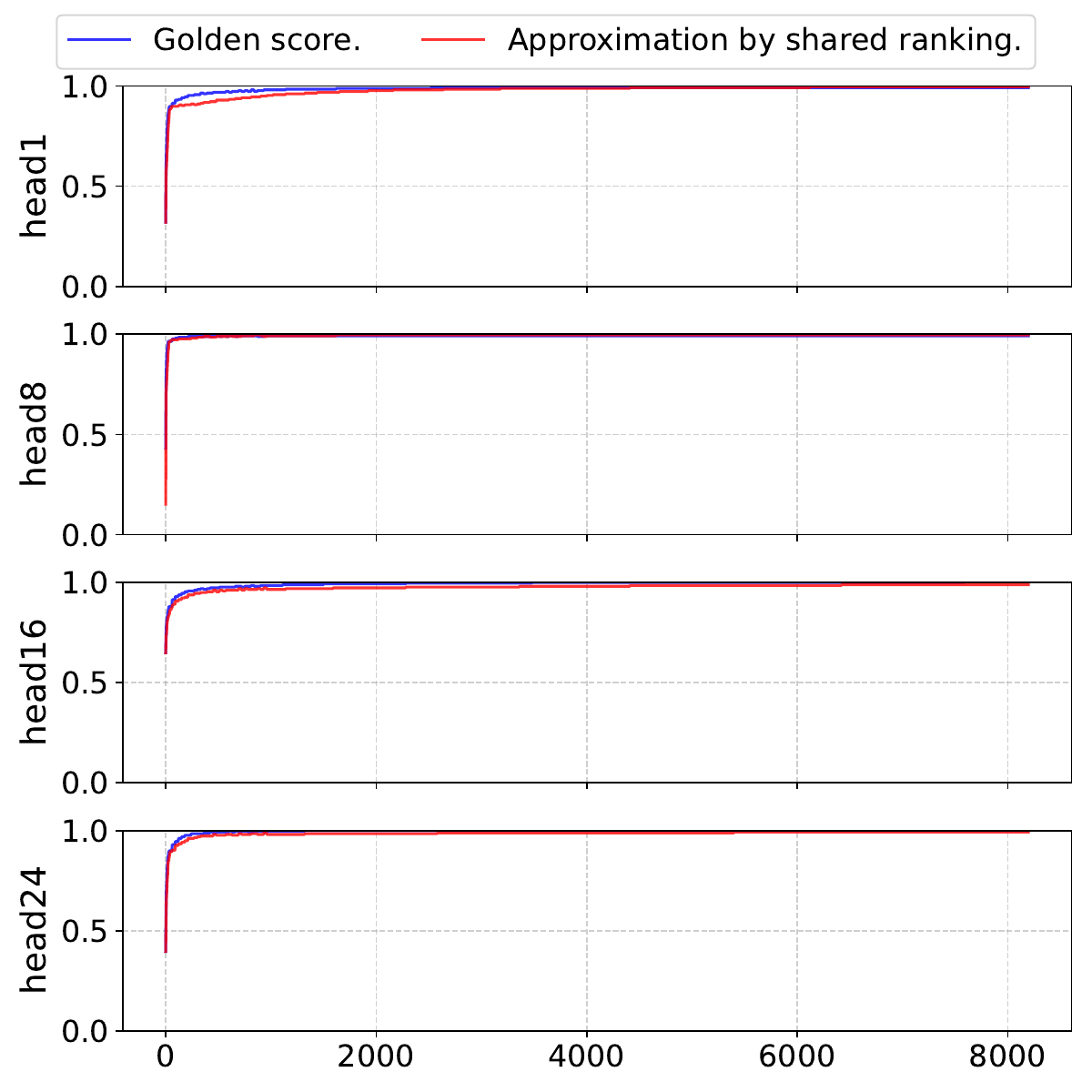}}
   \hfill
   \subcaptionbox{Layer 23
   \label{fig:fake:4}}
   {\includegraphics[width=0.24\textwidth]{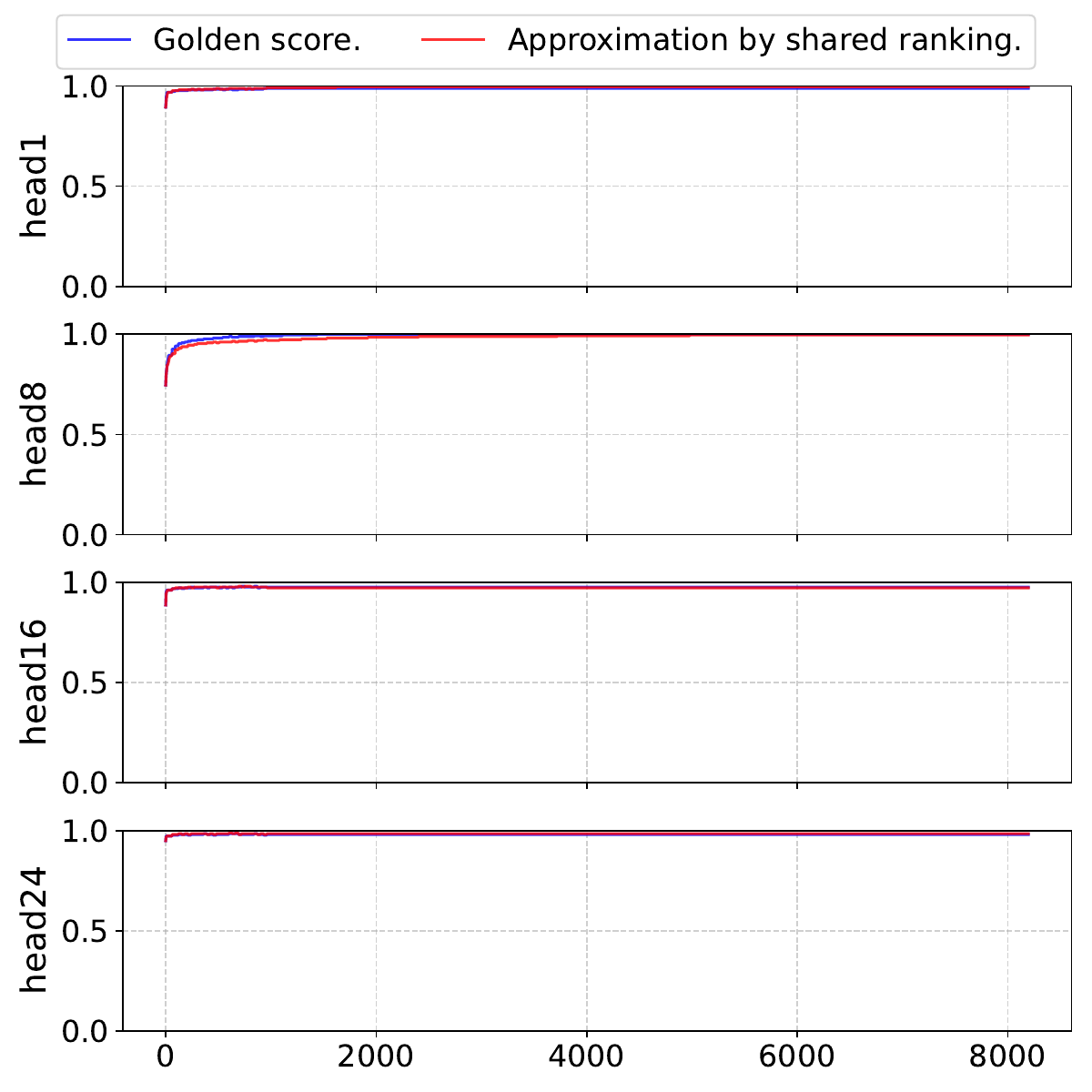}}
   \caption{Cumulative attention curves across different layers of Llama3.1-8B-Instruct on synthetic data
   from RULER.
   }
   \label{fig:fake}
\end{figure}

\begin{figure}[h]
   \centering
   \subcaptionbox{Layer 0.
   \label{fig:true:1}}
   {\includegraphics[width=0.24\textwidth]{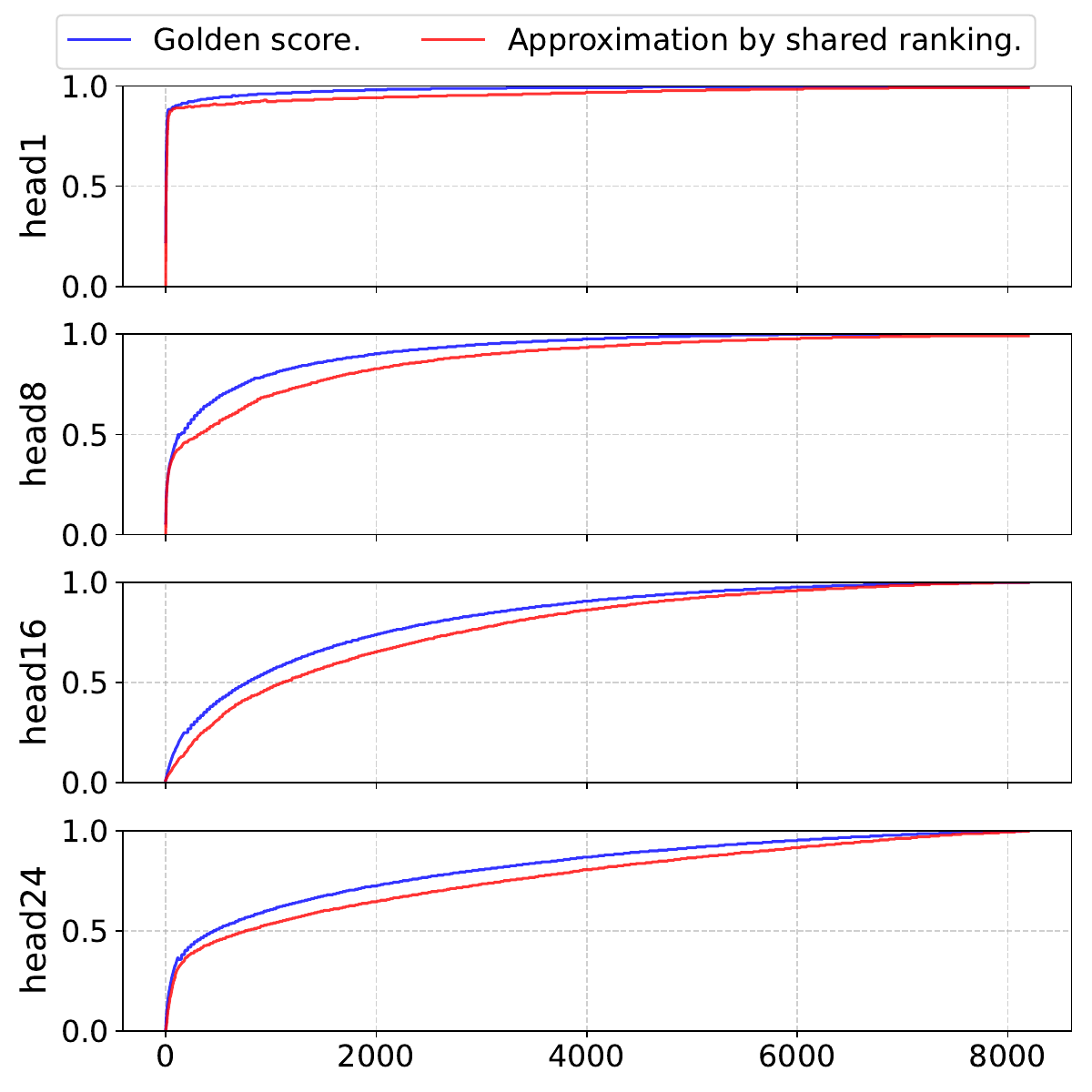}}
   \hfill
   \subcaptionbox{Layer 7.
   \label{fig:true:2}}
   {\includegraphics[width=0.24\textwidth]{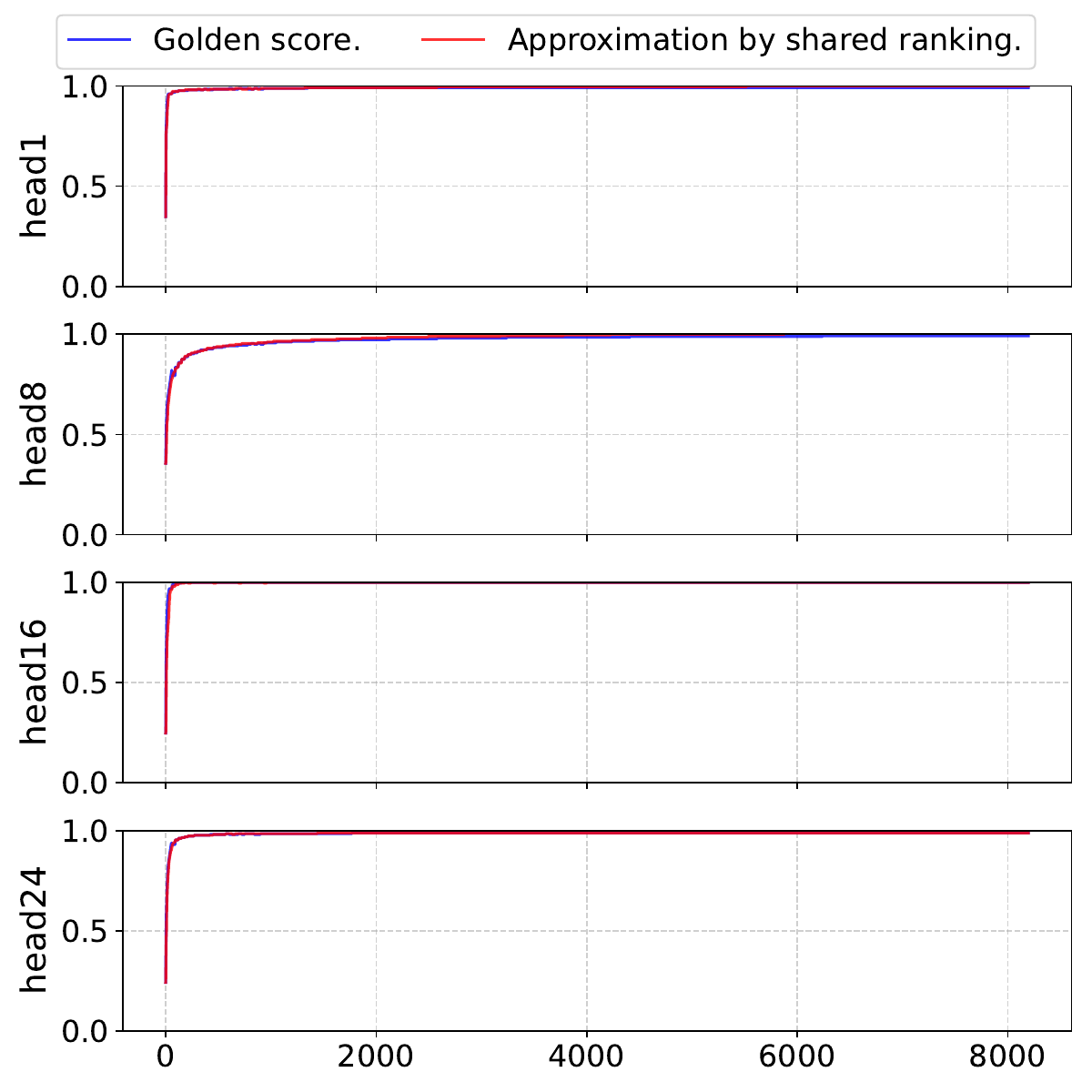}}
   \hfill
   \subcaptionbox{Layer 15.
   \label{fig:true:3}}
   {\includegraphics[width=0.24\textwidth]{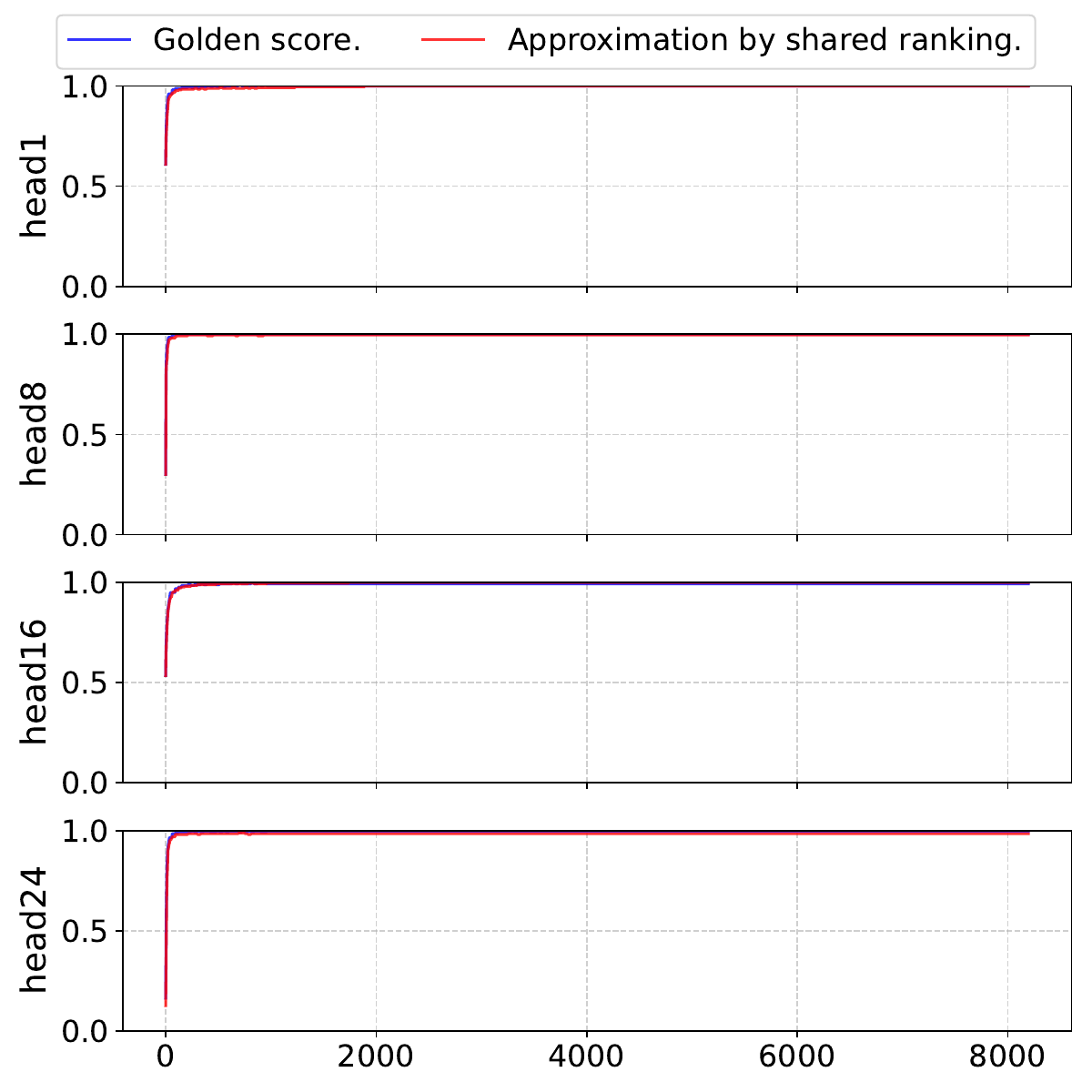}}
   \hfill
   \subcaptionbox{Layer 23.
   \label{fig:true:4}}
   {\includegraphics[width=0.24\textwidth]{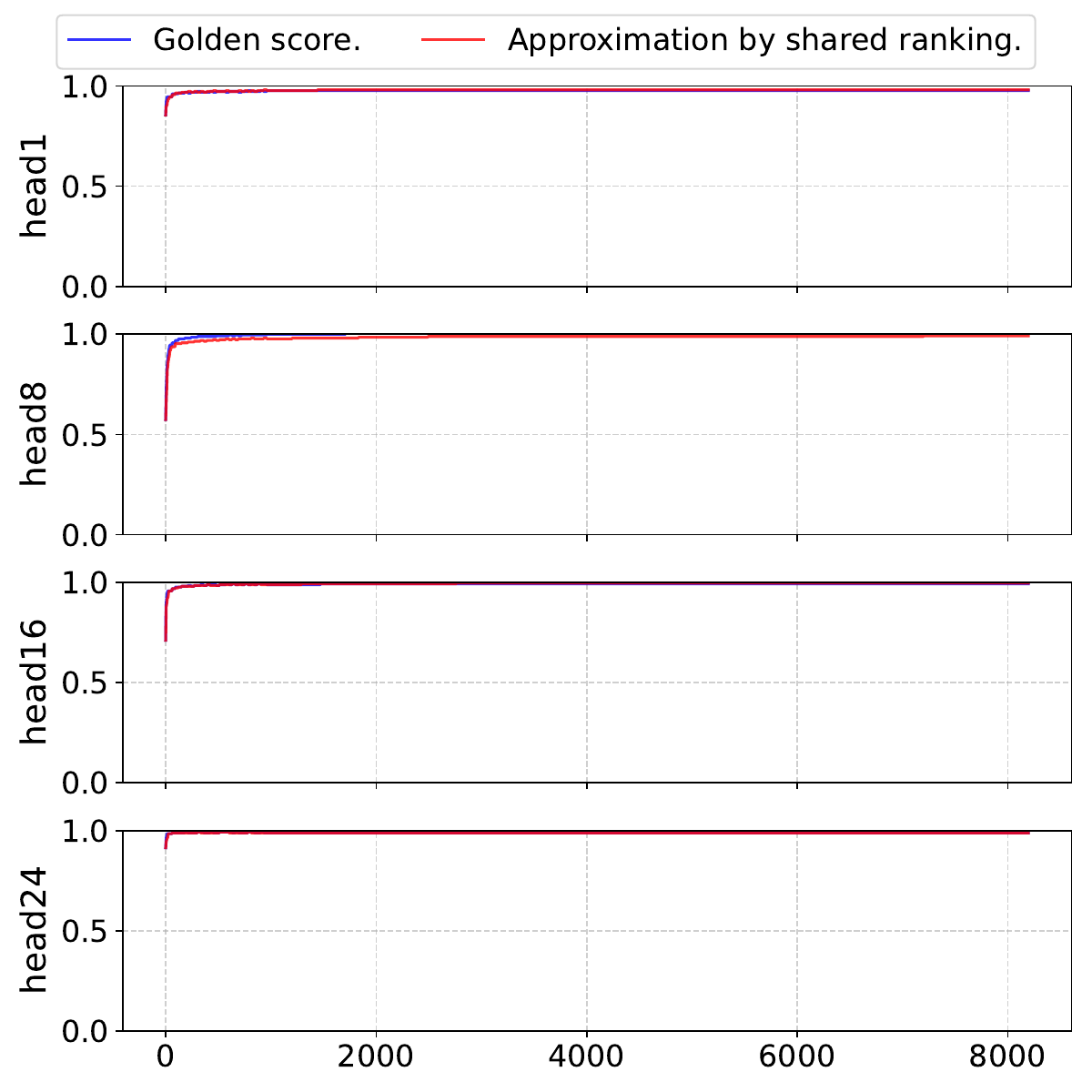}}
   \caption{Cumulative attention curves across different layers of Llama3.1-8B-Instruct on real-world data
   from LongBench v2.
   }
   \label{fig:true}
\end{figure}

% motivation
To verify the domain-specific differences of the phenomena observed in Section \ref{sec:finding},
we conducted observational experiments on both synthetic and real-world datasets.
% Figure
As shown in Figures \ref{fig:fake} and \ref{fig:true},
we observe the cumulative attention score curves for input tokens ranked by both the proxy head and the original heads
across different datasets.
% From
For synthetic data, we use RULER dataset samples consistent with \S \ref{sec:finding},
while real-world data are sampled from LongBench v2 \citep{bai2024longbench2}.
% Conclusion
It can be observed that the proxy head exhibits relatively consistent behavior across the two different domains, validating the generality of the proposed method.

\end{document}